\documentclass[final,5p,times,twocolumn]{elsarticle}

\usepackage{color}
\usepackage[table, dvipsnames, HTML]{xcolor}
\definecolor{light-gray}{gray}{0.95}
\definecolor{wheat}{HTML}{FFFF99}
\definecolor{header_blue}{HTML}{39a6F1}
\definecolor{gainsboro}{HTML}{E0E0E0}
\definecolor{safegray}{HTML}{333333}
\usepackage{subcaption}
\DeclareCaptionSubType*[arabic]{figure}
\usepackage{graphicx}
\usepackage{amsmath}
\usepackage{amssymb}  
\usepackage{multirow}
\usepackage{array}
\usepackage{flushend}
\usepackage{sidecap}
\usepackage[normalem]{ulem}

\usepackage[utf8]{inputenc}
\DeclareUnicodeCharacter{0252}{\"u}
\DeclareUnicodeCharacter{0246}{\"o}
\DeclareUnicodeCharacter{0237}{\'\i}
\DeclareUnicodeCharacter{0225}{\'a}

\usepackage{verbatim}
\usepackage{arydshln}
\usepackage[parfill]{parskip}
\usepackage[export]{adjustbox}

\def\*#1{\mathbf{#1}}

\usepackage[colorlinks=true, citecolor=blue]{hyperref}

\journal{Robots and Autonomous Systems}
\begin{document}

\begin{frontmatter}
\title{Invariant Feature Mappings for Generalizing Affordance Understanding Using Regularized Metric Learning}

\author{Martin Hjelm, Carl Henrik Ek, Renaud Detry, Danica Kragic}

\begin{abstract}

This paper presents an approach for learning invariant features for object affordance understanding. One of the major problems for a robotic agent acquiring a deeper understanding of affordances is finding sensory-grounded semantics. Being able to understand what in the representation of an object makes the object afford an action opens up for more efficient manipulation, interchange of objects that visually might not be similar, transfer learning, and robot to human communication. Our approach uses a metric learning algorithm that learns a feature transform that encourages objects that affords the same action to be close in the feature space. We regularize the learning, such that we penalize irrelevant features, allowing the agent to link what in the sensory input caused the object to afford the action. From this, we show how the agent can abstract the affordance and reason about the similarity between different affordances.

\end{abstract}


\end{frontmatter}

\section{Introduction}\label{introduction}

Abstraction is mankind's ability to condense and generalize previous
experience into symbolic entities that can act as proxies for reasoning
about the world. We ground these abstractions in our sensory experience
and let them take on different meanings depending on context
\cite{Barsalou:2008ff}. The task of learning to detect and abstract
affordances is different to visual categorization. A deeper
understanding is based on being able to detect intra-category
commonality rather than saliency.

\begin{figure}[ht!]
    \centering
    \includegraphics[width=.48\textwidth,clip,trim=55 338 630 170]{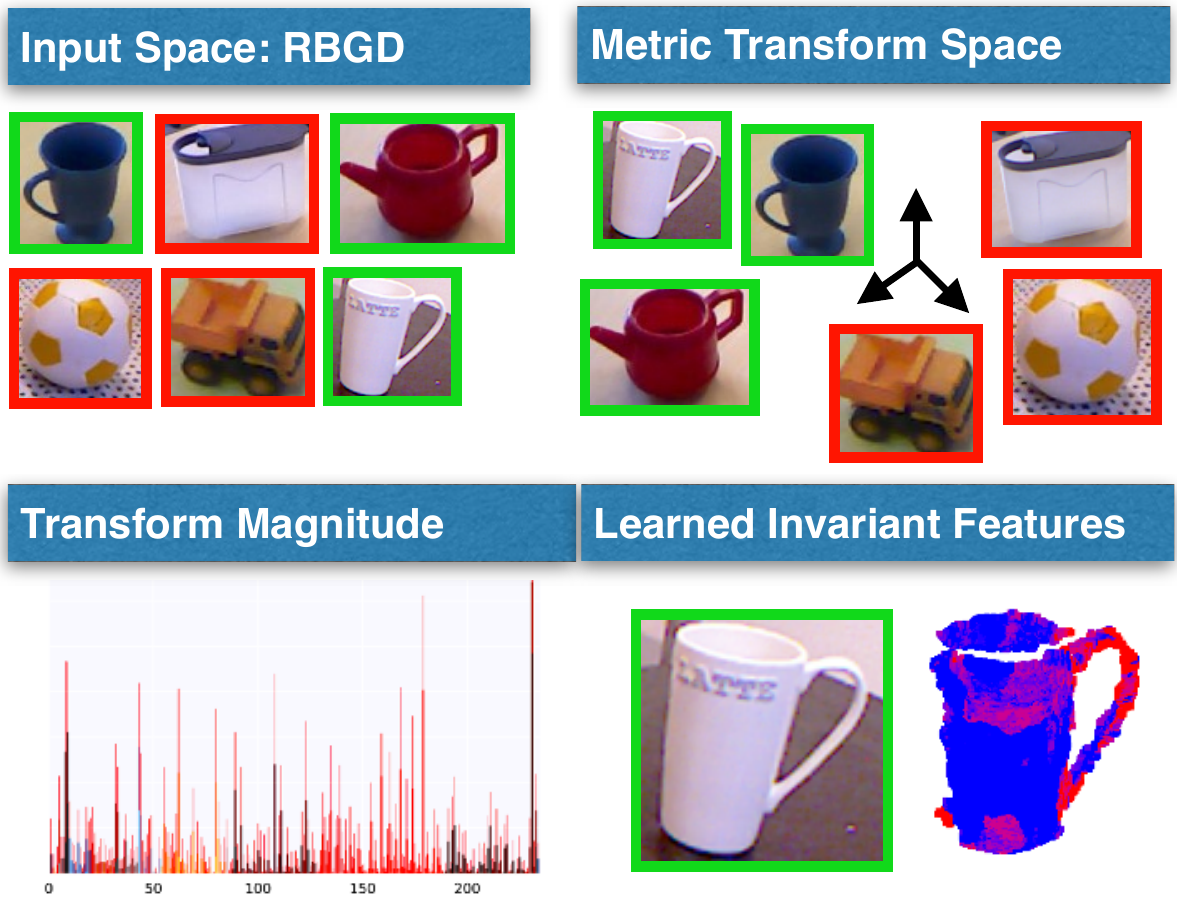}
    \caption{Figure summarizing our approach. For a specific affordance, we learn a linear transform, $\*L$, from a set of RGBD images of objects that both afford and does not afford an action. Each RGBD input is represented by a set of features $f(\*x)$. Under the transform items that afford the action are close in space and items that do not are far away. We regularize the learning of $\*L$ penalizing irrelevant features. We interpret the magnitude of the columns $\*L$ as feature selection enabling us to extract invariant features for the affordance. We map the invariant features onto the object thus finding important object parts for the affordance.}
\end{figure}

As an example of what this deeper understanding implies we can picture
an unstructured environment where the \emph{right} object for an
intended action is unavailable. An agent that can reason in an abstract
fashion can replace the unavailable object with other objects that
afford the same or similar actions. For example, it can replace a spoon
with a pen for stirring, replace a pan with a pot, etc. Hence categories
in this sense are not binary but loosely defined by a set of abstract
functional properties that makes up the common denominators of the
category.

An additional benefit of learning to abstract is that reasoning about
the similarity between categories becomes simpler as we are comparing
similarities across subsets of the feature space. The agent can thus
assemble hierarchies of clusters of similar actions, that in turn
enables reasoning within specific action domains, that in turn enables
better planning and synthesizes of explorative strategies in unknown
domains.

Having this cognitive ability is extremely useful. This paper thus
proposes a method for learning affordance abstractions, showing how the
agent can ground them in its own sensory input, and use it for reasoning
about the semantic similarity between objects.

We hypothesize that the abstract representation of an affordance
category is a latent space of the general space of vector representation
of objects. Associated with this latent space is a metric that we can as
a proxy for reasoning about similarity. We learn this similarity metric
from the data guided by the notion that similar items should be close in
the latent feature space and dissimilar items far away. In this paper,
this means learning a feature transform, $\*L$, that lets us reason in
the latent space under a Euclidian metric.

Analyzing $\*L$ an agent can learn the relevant features for
classifying to the affordance and enables it to ground the affordance in
the selected part of the sensory input. This grounding helps the robot
to locate affordance specific parts of the object from a set of general
global and local object features. Simply put, we learn the agent to
point out which parts of the feature representation of an object are
important for the affordance. By extension we enable the agent to point
out the physical parts of an object that is important for classifying it
to the affordance.

To learn to abstract affordance categories into a set of common features
the agent must ground the affordance in its own sensor input. The
grounding and abstraction allow for reasoning about the similarity
between affordances as it is reasonable to expect similar affordances to
have similar sets of common features. In the light of this, we propose a
novel interpretation: \emph{that we can understand the grounding of the
features for the categories through the similarity transform of the data
itself rather than through an analysis of data points in the latent
feature space}. We show that this semantic meta-similarity analysis is
possible through reasoning about the distance between the transforms,
$\*L$.

To form a complete understanding of an affordance the agent needs to
learn from interaction with the object, observing object, action, and
effect (OAE) triples. However, it has been argued that human design of
objects follows or should follow certain design principles that through
simple cues reveal the affordance of an object to the human observer
\cite{norman2002design}. In the light of this, this paper asks how much
of an affordance can we understand from just observing a set of objects
that afford the same action? What similarities exist in the feature
space of a category and can we deduce them just from observing category
and non-category members? Are these abstract similarities relevant for
interaction with the object?

We start by giving a wide perspective on current approaches to
affordance learning and go into detail on related work connected to the
proposed method. We proceed to describe our approach in detail and give
experiments showing how an agent can learn abstractions for affordance
categories and how it can reason about these categories. We end by
outlining some important principles and future work that needs
addressing.

\section{Perspective on Affordance
Learning}\label{perspective-on-affordance-learning}

This paper learns an agent to abstract affordance categories by
observing common features in the representation of objects that affords
the same action. Specifically, we learn a linear transform, $\*L$,
that gives us a latent representation where similar items close and
dissimilar items far away. The latent representation enables us to
compare items using the Euclidean metric as a proxy for similarity. We
penalize the learning of $\*L$ such that the transformation only
selects and transforms relevant features. We interpret the selected
features as an abstraction of the affordance we are learning and the
magnitude of $\*L$ as a measure of the relative relevance of a
feature. This relevance enables us to pinpoint important parts of the
objects belonging to an affordance category. Further on, we measure the
similarity between affordances by measuring the distance between the
transform magnitudes, that is, we are able to abstract affordance
categories and compare the abstractions.

Our approach to the affordance learning problem is thus quite different
to the general affordance learning research being done in robotics which
we divide into developmental methods based on exploration and methods
that learn to predict more advanced affordances from demonstration (LfD)
or annotated datasets.

\hypertarget{developmental-methods}{%
\subsection{Developmental Methods}\label{developmental-methods}}

Developmental methods have so far followed a paradigm of measuring
object, action, and effect (OAE) triples. They focus on simple
affordances such as pushing, rolling, simple tool use, etc., where the
outcome of an action is clear and measurable
\cite{Chao:2011id, Griffith:2010kw, Griffith:2009cm, Hogman:2013tg,Modayil:2008it, Montesano:2008dg, Niekum:us, Schenck:2013ts, Sinapov:2008dt, Stoytchev:2005il, Sahin:2007gr, Yuruten:2013hr, DBLP:conf/icra/DehbanJKS16,DBLP:conf/icdl-epirob/GoncalvesASJB14, DBLP:conf/icra/PenkovBR17, DBLP:journals/tamd/IvaldiNLDPFOS14, DBLP:conf/icra/MarTMN15, DBLP:conf/icvs/KraftDPBPK09}.
This is sensible since their objective is to learn a robot with limited
cognitive and motor abilities to connect OAE triples. Learning is often
unsupervised and explorative
\cite{DBLP:conf/icvs/KraftDPBPK09, Griffith:2009cm, Griffith:2010kw, Modayil:2008it}
and based on learning thresholds \cite{Chao:2011id,Niekum:us} for the
perceived features thus requiring clear pre- and post-conditions. These
threshold operations are similar in nature to abstracting the feature of
a category, however, they are often semi-automated and built using
heuristics rather than the automatic process of our approach.

One of the more complete models, with regard to structuring the learning
as well as showing experiments in real environments, comes from
\cite{Modayil:2007uu, Modayil:2008it}. The authors represent objects not
as physical entities but as a \emph{``hypothesized entity that accounts
for a spatiotemporally coherent cluster of sensory experience''}. They
represent objects by a set containing a tracker, percept, classes, and
actions, which are all more or less temporal. The most interesting
aspect of this formulation is the representation of objects as
consistent sensory inputs over time and associated with action
possibilities that produce certain outcomes. This more integrated view
of learning about objects and interacting with the world is much closer
to the idea of symbolic grounding and how some researcher thinks humans
organize grounded knowledge.

More recent affordance-based learning approaches also employ the OAE
paradigm still with simple actions but with some form of convolutional
neural network (CNN) used together with massive amounts of collected OAE
triples
\cite{DBLP:journals/corr/PintoGHPG16, DBLP:journals/corr/AgrawalNAML16, DBLP:conf/icra/PintoG17}.
In \cite{DBLP:journals/corr/AgrawalNAML16} the authors hypothesize that
humans have an internal physics model that allows them to understand and
predict OAEs. They suggest learning a similar model via a siamese CNN of
the image input from before and after an action.
\cite{DBLP:journals/corr/PintoGHPG16} takes a similar approach, however,
there the novelty lays in the construction of a branching deep net. The
network has a pre-trained common base that branches out with nets
pre-trained for inference of pinch grasping, pushing, and pulling
actions. The base net feeds its output into the branches and receives
feedback from them updating both weights in the base and branches. This
enables the net to refine the input to cater to specific tasks. This is
similar to the current perception of how humans process visual
information. The processing starts with a unified preprocessing of the
visual input and then branching into cortical areas that handle vision
for action and vision for cognition.

It is obvious that deep methods offer a great advantage in processing as
they can take raw image input and consistently produce good results. In
addition, they can process large amounts of training data that a robot
needs to learn affordances that are not toy examples. However, the
drawback of these methods is that deep nets are somewhat of black box
method lacking in interpretability, they are also data inefficient, and
can be unreliable in prediction. To the contrary our approach yields
interpretable results, it allows us to locate the position of important
features on the object for specific affordances, and to reason about the
similarity between affordances.

\hypertarget{advanced-affordances}{%
\subsection{Advanced Affordances}\label{advanced-affordances}}

Learning OAE triples are thus a seemingly agreed upon fundamental
component to learning affordances. However, learning affordances from
everyday object interactions is more complicated. Actions are complex.
They involve several steps of manipulation and outcomes are therefore
not always as clear-cut. Efforts so far have thus focused on some form
of supervision either in acquiring the training data, the provision of
labels, or implicitly in the model. A majority of the models tries to
infer the affordance or the action instead of learning the robot to
generalize, understand, and perform the action associated with the
affordance.

Most methods take a standard supervised computer vision approach, that
is, categorizing labeled images, sequences, or action commands
\cite{Fritz:2006kh, Stark2008, Montesano:2009wo, Hermans:2011vz, JieSun:2010kv, DBLP:conf/icra/YeYMFA17, Myers:2015fa, DBLP:conf/iros/NguyenKCT16a}.
Others try to model relationships between an observed actor, typically a
human, and the objects it interacts with
\cite{Pieropan:2013jm, Koppula:2012uy, Gall:2011us, Gall:2011us, Aksoy:2011wy, Kjellstrom:2011fh, Butepage:2018up}
learning affordances and actions jointly. However, robots are frequently
used as well and they are generally equipped with some form of
pre-programmed knowledge such as actions, action effects, features, or
object knowledge
\cite{Wang:2014hi, Thomaz:2009uk, Chu:2016:LOA:2906831.2906869, Montesano:2008dg},
to assist in the learning. These methods are good at what they do:
predicting actions and outcomes from visual input. However, for a robot
trying to understand the action and perhaps learn to perform the action
itself these methods describe discretization of sensory input and
knowledge from a human perspective, not from the robot's own sensory
perspective.

Our approach is similar in that we learn from labeled images, but with
multiple affordance labels for the whole image as in
\cite{Hjelm:2015hw, Nagarajan:2018uf} instead of learning to predict
pixelwise labels in
\cite{Myers:2015fa, DBLP:conf/iros/NguyenKCT16a, DBLP:conf/icra/DoN018, DBLP:conf/iros/AbelhaG17,DBLP:conf/iros/NguyenKCT17, detry2017b}
and without the addition of actions and outcomes. Our goal is to ground
the affordance in the representation of the object. As stated in the
introduction our interest lays in what kind of abstractions an agent can
learn from observing the common features in a category and how we can
use these grounded features to reason about and perform the affordance.

\hypertarget{attribute-learning}{%
\subsection{Attribute Learning}\label{attribute-learning}}

Humans use rule-based and similarity reasoning to transfer knowledge
about categories but it is almost certainly not how our visual system
categorize everyday objects at the basic category level. Nevertheless,
works exploring classification by attributes or attribute learning are
important because they touch on the deeper question of how to learn the
invariant features of categories, albeit from high-level abstractions.
This is an extremely important ability to have when generalizing
affordances. When humans substitute objects, it tends to happen in an
ad-hoc fashion. We base the selection process on similarity comparisons
across the abstraction we have for an affordance to motivate the
substitution.

These types of attribute approaches have mostly been explored in
computer vision
\cite{Ferrari07, Lampert:2009te, Farhadi:2009dn, Malisiewicz:2008cy}.
\cite{Ferrari07} segments images and learns a graphical model over the
segments that models relations between segments and contexts enabling it
to predict patterns such as striped, colors, etc. \cite{Lampert:2009te}
associate specific attributes with specific image categories such that
they can infer the image class from knowledge about the attributes. The
attributes act as an intermediate layer in a graphical model which
enables conditioning novel classes on learned classes and attributes.
The model does not recognize new attributes but rather rely on the
notion that learned attributes contain information relevant for novel
classes.

The approach most similar to ours is that of
\cite{Farhadi:2009dn, Malisiewicz:2008cy}. \cite{Malisiewicz:2008cy}
formulate the categorization problem as data association problem, that
is, they define an exemplar by a small set of visually similar objects
each with associated distance functions learned from the data.
\cite{Farhadi:2009dn} approach equates the ability of attribute
prediction with the ability to predict image classes from the learned
attributes. They stack a broad number of different features and use
feature selection to filter out irrelevant features. They realize that
the number of attributes they have specified is not sufficient to
classify to the specific categories and opt to learn additional
attributes from the data.

The abstractions we want to learn can also be considered discriminative
attributes, however, we learn these through similarity comparisons
rather than through discrimination. Our aim is to simultaneously learn
to predict categories and abstract them as we view them as different
aspects of the same process.

Robotics has also explored the attribute-based inference approach.
\cite{Hermans:2011vz, JieSun:2010kv} uses a Bayesian Network (BN) that
relates class, features, and attributes. The authors learn a robot to
recognize key attributes of objects such as size, shape, color,
material, and weight, which they use to predict affordances such as
traverse, move, etc. They compare their approach to affordance
prediction with an SVM trained directly on the feature space. The direct
approach performs comparably or better than the attribute-based
approach, the explanation they propose is that the feature space
contains information not directly explainable as any specific semantic
attribute.

This serves to illustrate that we should not program autonomous robots
in unstructured environments to process the world in hierarchies of
abstract symbols derived from our own human sensorimotor systems
\cite{Brooks:1990uu}. Human language is abstract semantic symbols
grounded in invariant features to make conveying, planning, and
reasoning easy. All psychophysical evidence points to that human's
sensorimotor systems do not count attributes to recognize an object.
Further on, we cannot expect different sensorimotor systems, humans, and
robots, to produce the same semantic grounding unless they are exactly
identical in construction and experience. This is a key feature of our
work. We learn the invariant features from detecting similarities in the
representation of intra-category objects. The agent is thus able to
ground the semantic meaning of the affordance in its own sensory-motor
system and enables it to locate physical parts of the objects that are
important for classifying to the affordance. However, we never use these
parts to infer if an object affords an action or not.

\hypertarget{similarity}{%
\subsection{Similarity}\label{similarity}}

Measuring similarity is difficult especially for high-dimensional
representations as any arbitrary measure would have to treat each
dimension as equally valuable. This will leave the important features
open to being drowned out by either noise or the amount of non-relevant
features. Further on, different metrics are useful for certain
distributions of the data while being detrimental for others.

One way of solving this involves specifying a relevant representation or
measure for each category. However, this solution does not scale and
contrasts with the idea that an agent should ground semantics in its own
sensory input. The other approach, which we adhere to, is learning the
similarity from the data. We consider the useful representation for a
category to be a latent representation of a more general object
representation. Learning the latent representation, in turn, enables us
to use the Euclidean metric for reasoning about similarity.

The similarity measures used in affordance learning are mostly used to
describe the similarity between OAE triples or a subset of them. Many
formulate their own measures or use the standard Euclidean measure
\cite{ Gall:2011us, Aksoy:2011wy, Griffith:2010kw, Griffith:2009cm, Modayil:2008it, Schenck:2013ts, Sinapov:2008dt, Stark2008, Wang:2014hi}.
The measures are often used in an unsupervised setting to cluster for
affordance categories. Other use kernels as an implicit measure of
similarity in supervised learning
\cite{Ek:2010wk, Hermans:2011vz, Montesano:2008dg, Pieropan:2013jm, Yuruten:2013hr}.

Entropy \cite{Shannon:2013iy} is sometimes used to compute distances
between distributions that describe possible actions or object
categories \cite{Hogman:2013tg} or measure the stability of unsupervised
category learning \cite{Schenck:2013ts, Gall:2011us, Griffith:2010kw}.
Lastly, another popular approach is to model associations as graphical
models, Bayesian Networks (BN) or Conditional Random Fields (CRF), as
they are good at describing the temporal nature of object interaction or
other complex associations
\cite{Ek:2010wk, Kjellstrom:2011fh, Koppula:2012uy}. Here probability
becomes a proxy for similarity.

To the best of our knowledge, no previous method has approached the
affordance classification problem by learning the metric from the data.
We can think of the CNN approaches described above as learning a
transform that enables an implicit similarity mapping, however, as
opposed to our approach they are unable to locate what in the input
caused the classification. Further on, CNN based approaches projects
non-linearly onto massively high-dimensional spaces using massive
amounts of data. We instead show that our linear projection can reduce
the dimensionality down from 322 dimensions to 3 with no significant
loss in accuracy using low amounts of data. At same the time, our
sparsity-inducing regularization forces the projection to only use a
small subset of the features, on average \(30\%\) of the feature space.
Finally, we learn what physical part of the objects in a category are
relevant for the category giving the agent a deeper connection between
sensory input and the actionable parts of the objects.

In this sense, the most similar approach to ours in learning the feature
space is a method by \cite{Sun:2013hq}. They learn a feature codebook
over the RGBD space of objects by optimizing towards a compact
representation of the feature space in an unsupervised fashion similar
to an autoencoder. The authors use the codebook to find a lower
dimensional representation of objects and to classify object attributes.
They show that by regularizing the classifier that they can learn which
codewords are important for specific attributes. However, this approach
is computationally taxing as they learn the codebook and the latent
representation simultaneously. Contrary to our approach they aim for a
general representation for all tasks rather than utilizing class labels
to learn task-specific representations.

\section{Methodology}\label{methodology}

Our goal is to learn a feature transform, $\*L$, for each affordance,
that given a general object representation, $\*x$, outputs a latent
representation, $\*z$. This latent representation should have the
quality that objects that affords an action should be close in the
latent space and others far away. This implies that we can use the
Euclidean metric as a proxy for measuring similarity. We learn $\*L$
from a set of \(n\) input-target pairs, \(\{\*x_i, y_i\}\). Here $\*x$
is a general feature vector where \(y\) is a label denoting if the
object affords an action or not.

Given a set of feature transforms $\*L$ for different affordances our
approach has four goals. We want to:

\begin{itemize}
\item
  Learn what features of the general objection representation $\*x$
  are important for classifying instances to each of the affordances.
\item
  Formulate a general abstract representation of the affordance-based
  upon the relevant features.
\item
  Given an object locate the relevant parts on the object.
\item
  Model the relationship between the affordances such that we can
  understand which affordances are similar.
\end{itemize}

\hypertarget{features}{%
\subsection{Features}\label{features}}

We capture objects as an RGBD image using a Kinect sensor and convert it
into a 2D image and a point cloud. The point cloud representation is
noisy and many times parts of the object are missing due to the
reflective material of those parts. We can, therefore, expect the
relevance and reliability of the features to vary substantially across
the different affordance categories. Further on, it is difficult to know
what features are important for classifying to a specific affordance.
Because of this, we choose to stack a number of global and local
features and let the algorithm decide on relevance. The stacking gives a
feature vector dimension of \(234\).

The global features are:

\begin{itemize}
\item
  \textbf{Object volume} - the volume of the convex hull enclosing the
  object point cloud.
\item
  \textbf{Shape primitive} - similarity to primitive shapes cylindrical,
  spheric, and cubic as fitted by the RANSAC algorithm.
\item
  \textbf{Elongation} - the ratio of the minor axes to the major axis.
\item
  \textbf{Dimensions} - the length of the sides of the object.
\item
  \textbf{Material} - Objects often consists of different materials. We
  want a vector representation that gives a score for the different
  materials of an object. To finds these scores we train the SVM to
  classify textures \emph{glass, carton, porcelain, metal, plastic,
  wood}. The input is the concatenation of a Fisher Vector (FV)
  representation of the SIFT features of the image and the output of the
  5th layer of a re-trained GoogleLeNet. We take the scores of the SVM
  over an object as the decomposition score of the different materials.
\end{itemize}

We motivate these global features by research showing their usefulness
in predicting variables involved in human grasping and affordances,
e.g.~\cite{Baugh1262, Buckingham:2009dd, Fabbri:2016bd, Feix:2014cy, Martin:2007ck, Grafton:2010jo, Jenmalm:2000ui, 10.1371/journal.pone.0025203, 10.1371/journal.pone.0025203}.

The local features are:

\begin{itemize}
\item
  \textbf{Image gradients} - histograms of intensity and gradient order
  1, 2, 3.
\item
  \textbf{Color quantization} - the mapping of colors to a finite set of
  colors and computing the histogram over the mapped colors.
\item
  \textbf{Color stats} - entropy, mean, and variance over the color
  quantized object.
\item
  \textbf{FPHF} - Bag-of-Words over Fast Point Feature Histograms
  \cite{Rusu:2009hf} for a number of radius scales.
\item
  \textbf{HoG} - Bag-of-Words representation over the HoG
  \cite{Dalal:2005to} features of the image.
\end{itemize}

Again we motivate these features by studies showing their usefulness,
especially shape descriptors
e.g.~\cite{Kruger:2013wg, norman2002design}. Due to the point cloud
representation, we only need to keep the portion of features associated
with the point cloud. For example, for the gradients, we only compute
the gradients for pixels associated with the point cloud. This works for
all features except for HoG as it uses patches overlapping the image.

\hypertarget{learning}{%
\subsection{Learning}\label{learning}}

As discussed in the introduction we use distance as a proxy for
similarity. However, with low amounts of data, it is difficult to
construct a general high-dimensional feature space that works well under
some metric for a number of different labelings.

We, therefore, want to learn a transform, $\*L$, for each affordance
that puts similar instances close in space and dissimilar instances far
way. The transform should help us locate parts of the feature space that
is relevant and project onto a subspace, \(d\), such that alleviates the
curse of dimensionality.

\begin{figure}
    \centering
    \includegraphics[width=.7\columnwidth,clip,trim=240 370 390 110]{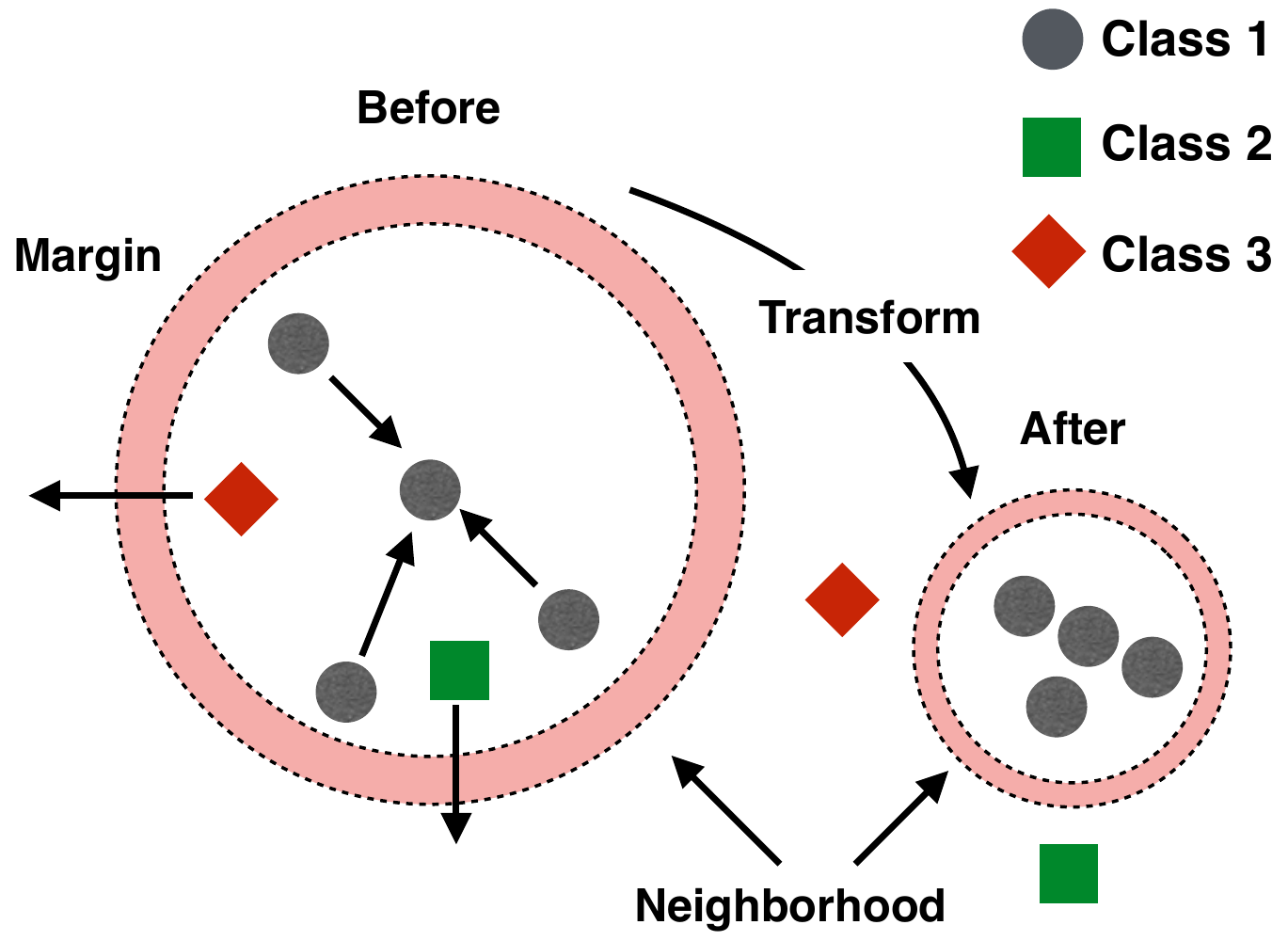}
    \caption{Figuratively LMCA optimization first finds the $k$ neighbors, called target neighbors, for each instance by evaluating the nearest neighbor in the untransformed space. The algorithms then try to find a transform, $\*L$, that pushes those target neighbors towards the instance, the neighborhood center, while at the same time pushing non-class members out of the perimeter of the neighborhood.}
    \label{fig:chp5:mlearning_explanation}
\end{figure}

To this end, we use a regularized version of the Large Margin Component
Analysis (LMCA) metric learning algorithm \cite{Torresani:2006wb} which
we will refer to as LMCA-R. LMCA learns a linear transformation,
$\*L$, of the input space that pushes the \(k\) class
nearest-neighbors (NN) of every instance closer together while pushing
non-class members outside a margin as illustrated in
fig.\ref{fig:chp5:mlearning_explanation}. We learn $\*L$ using
gradient descent over the following loss function,

\begin{equation}
\label{eq:LMCAloss}
    \begin{aligned}
    & \epsilon(\mathbf{L}) = \sum\limits_{i,i \leadsto j} w_i ||\mathbf{L} (x_i-x_j)||^{2}  \\
    &  +\! c\!\!\! \sum\limits_{i,i \leadsto j,l} \!\!\! w_i \; y_{il}\;h(||\mathbf{L} (x_i-x_j)||^{2} \!-\! || \mathbf{L} (x_i-x_l)||^{2} +1 ) \\
    & +\! \lambda \sum_{j=1}^{D} \left\lVert L_j  \right\rVert_{2}.
    \end{aligned}
\end{equation}

Here \(i \leadsto j\) means the \(k\) nearest neighbors of the instance
\(i\) that belong to the same class, \(y_{il}\) is a binary variable
that is zero if \(i\) and \(l\) have the same label, and one otherwise.
The first term penalizes large distances for the \(k\) NN and the second
term penalizes non-class instances that are closer to the instance
\(x_i\) than the class \(k\) NN, by a margin of \(1\). \(c\) is a
constant controlling the relative importance of the pushing component
and \(h\) is the differentiable smooth hinge loss \cite{Rennie:2005ds}.

\(w_i\) is a weight term that aims to balance the learning in terms of
false-positive rates when some of the classes have few numbers of
exemplars. We formulate the weights as \(w_i = \frac{N}{N_i}\) where
\(N\) is the total number of data points and \(N_i\) is the number of
instances in the class that \(i\)-th instance belongs to. The
justification here is that if we assume that each of the summands, in
the loss function, are roughly similar in magnitude then the weight
factor will level the contribution from each class to the loss. For
example,

\begin{equation}
    \frac{N}{N_1} \sum_{C_1} x_i + \frac{N}{N_2} \sum_{C_2} x_i
    \approx \frac{N}{N_1} N_1 x_i + \frac{N}{N_2} N_2 x_i
    = N x_i + N x_i.
\end{equation}

The reason for multiplying by \(N\) is to keep the ratio at a reasonable
value to avoid numerical instability.

The last term in eq.\ref{eq:LMCAloss} is a penalization term due to
\cite{Obozinski:2009fm}. It is the sum of the \(l_{1/2}\)-norm applied
to the columns of $\*L$. The \(l_{1/2}\)-norm is simply the \(l_1\)
norm, of the \(l_2\) norm, of the columns of the, transformation matrix
$\*L$. The \(l_2\)-norm is the crucial factor as it helps contain the
full column, reducing it fully. This means that it will remove
irrelevant features completely instead of zeroing individual matrix
elements of $\*L$ as happens with the \(l_{1}\)-norm over the matrix
elements. \(\lambda\) is a constant controlling how much weight we want
to put on penalizing non-zero columns.

As $\*L$ is a projection we can choose to let it project onto a
subdimension, \(d\), which can be much less than \(D\). In the
experiments section, we show that we can project from \(234\) dimensions
down to \(3\) without a significant loss in accuracy, a reduction in
dimension of roughly \(99\%\).

\hypertarget{classification-and-analysis}{%
\subsection{Classification And
Analysis}\label{classification-and-analysis}}

\label{subsec:inv_classification_analysis} To classify to an affordance
category we formulate the problem as a binary decision problem, that is,
we learn a specific $\*L$ for each affordance class. We apply $\*L$
to the data and classify to the affordance using kNN where \(k\) is
equal to the number of neighbors used in the learning phase. Our use of
kNN is thus a direct evaluation of the metric.

We analyze the feature selection by taking the magnitude of the $\*L$
columns. Low values will mean an irrelevant feature while high will mean
relevant. To analyze the similarity between different affordances we
treat the magnitude vector as having a Gaussian multivariate
distribution and use the KL-divergence as the distance measure.

The affordance learning problem is a multiclass problem. However, two
factors motivated us to switch from multiclass to a binary decision
problem. Firstly, for the feature selection analysis to work, we need to
pit the objects in one affordance category against a wide range of
different objects. If we are learning multiple classes simultaneous this
analysis is not possible; the feature selection will instead show good
general features. Secondly, learning multiple classes at the same time
is not optimal as we would use fewer parameters and data points for each
problem.

\section{Experiments}
We motivate our experiments by the following three questions:

1. Does our approach select features that are sensible as an abstraction for explaining an affordance? 
2. Do the selected features map out a similar set of parts on all the objects in an affordance category?
3. How do the affordances relate to each other? Are the affordances we as humans view as similar equal to what the model deems as similar?
    
\subsection{Dataset}
We collected 265 RGBD images of everyday objects ranging from cups to cereal boxes, tools, cans, and water bottles. To collect the images we placed the objects on different flat surfaces and recorded an RGBD image using a Kinect camera. We took each image under different light conditions and tried to vary the pose of the objects to a reasonable amount. Many of the images had small parts or parts made of glass or metal leaving large holes in the depth recordings. Since each image is devoid of clutter it is simple to segment out the object by simply removing all point cloud points, not above the planar surface. 

We labeled each object as a binary vector specifying if it affords each one of the affordances in table \ref{table:affordance-classification}. Many of these affordances are quite vague and labeling is not as binary as in standard image classification. This vagueness follows from the vagueness in the definition of the affordance concept. Many objects that afford an action will under normal circumstances not be used for the affordance if other suitable objects are available.

\begin{table}[ht]
    \setlength\tabcolsep{3pt}
    \resizebox{1.0\linewidth}{!}{%
    \begin{tabular}{ l l l l l }
    \textbf{Affordance} & \textbf{kNN} & \textbf{LMCA-R} & \textbf{SVM} & \textbf{CNN-SVM} \\ \arrayrulecolor{safegray}\hline
    \rowcolor{gainsboro}
    \textbf{Containing}\;(124,142)  & $0.87\;(87.6)$ & $0.91\;(91.9)\;42\%$ & $\mathbf{0.92}\;(93.0)$ & $0.9\;(91.2)$\\
    \textbf{Cutting}\;(11,255)  & $0.36\;(95.2)$ & $\mathbf{0.43}\;(95.7)\;19\%$ & $0.22\;(94.9)$ & $0.39\;(92.6)$\\
    \rowcolor{gainsboro}
    \textbf{Drinking}\;(36,230)  & $0.64\;(90.4)$ & $\mathbf{0.79}\;(94.1)\;24\%$ & $0.71\;(91.6)$ & $0.68\;(88.4)$\\
    \textbf{Eating From}\;(25,241)  & $0.67\;(94.0)$ & $0.6\;(93.2)\;27\%$ & $0.59\;(92.5)$ & $\mathbf{0.71}\;(92.0)$\\
    \rowcolor{gainsboro}
    \textbf{Hammering}\;(21,245)  & $0.32\;(92.3)$ & $\mathbf{0.48}\;(93.0)\;34\%$ & $0.31\;(92.9)$ & $0.41\;(82.0)$\\
    \textbf{Handle Grasp}\;(56,210)  & $0.77\;(91.4)$ & $\mathbf{0.85}\;(94.0)\;36\%$ & $0.83\;(92.8)$ & $0.8\;(89.0)$\\
    \rowcolor{gainsboro}
    \textbf{Hanging}\;(45,221)  & $0.37\;(79.8)$ & $0.49\;(82.1)\;47\%$ & $0.07\;(81.6)$ & $\mathbf{0.59}\;(82.3)$\\
    \textbf{Lifting Top}\;(79,187)  & $0.63\;(77.3)$ & $0.63\;(79.0)\;45\%$ & $0.69\;(80.0)$ & $\mathbf{0.75}\;(86.3)$\\
    \rowcolor{gainsboro}
    \textbf{Loop Grasp}\;(31,235) & $0.38\;(87.0)$ & $0.45\;(86.6)\;39\%$ & $0.0\;(87.6)$ & $\mathbf{0.65}\;(89.4)$\\
    \textbf{Opening}\;(118,148)  & $0.86\;(87.1)$ & $\mathbf{0.89}\;(90.0)\;48\%$ & $\mathbf{0.89}\;(90.4)$ & $0.88\;(89.8)$\\
    \rowcolor{gainsboro}
    \textbf{Playing}\;(16,250)  & $0.51\;(96.0)$ & $\mathbf{0.64}\;(96.4)\;29\%$ & $0.62\;(96.2)$ & $0.45\;(85.7)$\\
    \textbf{Pounding}\;(86,180)  & $0.68\;(78.1)$ & $\mathbf{0.78}\;(85.8)\;46\%$ & $0.75\;(82.4)$  & $0.71\;(78.8)$\\
    \rowcolor{gainsboro}
    \textbf{Pouring}\;(162,104)  & $0.88\;(84.4)$ & $\mathbf{0.9}\;(87.9)\;49\%$ & $\mathbf{0.9}\;(87.6)$ & $\mathbf{0.9}\;(87.6)$\\
    \textbf{Putting}\;(56,210)  & $0.73\;(88.8)$ & $\mathbf{0.83}\;(92.6)\;33\%$ & $0.79\;(90.2)$ & $0.58\;(67.4)$\\
    \rowcolor{gainsboro}
    \textbf{Rolling}\;(105,161)  & $\mathbf{0.79}\;(83.2)$ & $0.78\;(83.1)\;53\%$ & $0.78\;(82.6)$ & $0.72\;(78.8)$\\
    \textbf{Scraping}\;(41,225)  & $0.77\;(93.4)$ & $0.78\;(93.6)\;37\%$ & $\mathbf{0.79}\;(93.4)$ & $0.75\;(90.6)$\\
    \rowcolor{gainsboro}
    \textbf{Shaking}\;(127,139)  & $0.86\;(86.4)$ & $0.9\;(90.8)\;46\%$ & $\mathbf{0.91}\;(91.5)$  & $0.89\;(90.0)$\\
    \textbf{Spraying}\;(9,257)  & $0.07\;(96.3)$ & $0.33\;(96.4)\;30\%$ & $0.05\;(96.2)$ & $\mathbf{0.56}\;(94.4)$\\
    \rowcolor{gainsboro}
    \textbf{Squeezing}\;(89,177)  & $0.6\;(73.0)$ & $0.68\;(78.5)\;50\%$ & $0.66\;(76.4)$ & $\mathbf{0.73}\;(80.5)$\\
    \textbf{Squeezing Out}\;(14,252)  & $0.37\;(95.8)$ & $0.45\;(95.5)\;36\%$ & $\mathbf{0.46}\;(95.8)$  & $0.34\;(92.3)$\\
    \rowcolor{gainsboro}
    \textbf{Stacking}\;(38,228)  & $0.78\;(94.6)$ & $\mathbf{0.81}\;(95.2)\;21\%$ & $0.72\;(93.0)$  & $0.76\;(93.4)$\\
    \textbf{Stirring}\;(39,227)  & $0.7\;(92.5)$ & $\mathbf{0.86}\;(96.0)\;25\%$ & $0.8\;(94.1)$ & $0.75\;(91.0)$\\
    \rowcolor{gainsboro}
    \textbf{Tool}\;(53,213)  & $0.84\;(94.1)$ & $\mathbf{0.91}\;(96.5)\;28\%$ & $0.89\;(95.7)$ & $0.88\;(95.4)$\\ \arrayrulecolor{safegray}\hline
    \textbf{Average} & $0.63\;(88.63)$ & $\mathbf{0.7}\;(90.75)\;37\%$ & $0.63\;(90.1)$ & $0.69\;(87.34)$  \\
    \end{tabular}}
\caption{Affordance classification 2 for kNN, regularized LMCA, and linear SVM. F1-score and accuracy in parenthesis, bold indicates the best value. The number of instances per class is given next to the task name, positive instances first. Most of the classes are unbalanced, giving F1-scores that are quite low the tasks that are highly unbalanced even though the accuracy is high, which is natural in many unbalanced binary classification tasks. LMCA outperforms both kNN and SVM both for F1-score and accuracy for most of categories and overall. In addition, LMCA discards on average roughly $60\%$ of the feature dimensions.}
\label{table:affordance-classification}
\end{table}

\subsection{Classification}
A prerequisite for answering the above questions is to first validate if the algorithm and features provide good affordance classification accuracy, that is, if the similarity metric we learn produces valid results. We compare our results to a kNN, and a linear SVM trained on the provided features. We also compare to an SVM trained on the output from the last fully connected layer of a pre-trained CNN as it has proven to be a good baseline.

As a pre-processing step, we standardize all data. We use five-fold cross-validation to learn the optimal parameters. For the kNN and SVMs, we also cross-validate against a PCA projection between $0-min(dim(X),20)$ dimensions where $0$ means no PCA is performed. For LMCA-R, we cross-validate for the impostor loss parameter, $c$, and the regularization parameter, $\lambda$.

We set the NN to $3$ for kNN and LMCA-R. For the SVMs we use the Scikit-learn library \cite{Pedregosa:2011tv} which uses LibSVM. We use a linear kernel and one-against-all classification. For LMCA-R we set the dimensionality reduction to $3$. For the CNN features we us Caffe \cite{Jia:2014:CCA:2647868.2654889} with a GooleLeNet model pre-trained on Imagenet. We extract the fifth layer giving us a $4096$ dimensional feature vector. 

We create $25$, $70/30$ training-test splits of the dataset. We give the results as averages over the $25$ splits in table \ref{table:affordance-classification}. We measure performance using the F1-score as our main metric as we are performing binary classification over many unbalanced categories. For example, for the spraying affordance, the accuracy is around $96\%$ but the best F1-score is $ 0.56$. As we can see from table \ref{table:affordance-classification} that LMCA-R performs best in a majority of the cases, outperforming kNN in all but one case.

Comparing the CNN features to the constructed we see that they perform roughly the same but for a few were one or the other significantly outperforms the other. It is difficult to pinpoint exactly why this is. We hypothesize that some affordances contain objects for which the depth recordings contain a high amount of noise. This propagates into uncertainties for the constructed features which are mostly dependent on depth recordings. For example, objects affording \textrm{hanging} usually have an arched part which can be difficult to record with sufficient accuracy as they are usually around 1 cm in diameter and thus close to the Kinect noise threshold. The CNN features, on the other hand, does not rely on depth measurements and are thus free of this constraint. We also see that the LMCA-R performs decently for these classes. This is due to the reweighing factor and the penalization that is able to disregard irrelevant features and weigh the lesser class as equally important.

\begin{figure}[!ht]
    \begin{subfigure}{0.48\textwidth}
        \includegraphics[width=\textwidth, clip, trim=7 8 7 6]{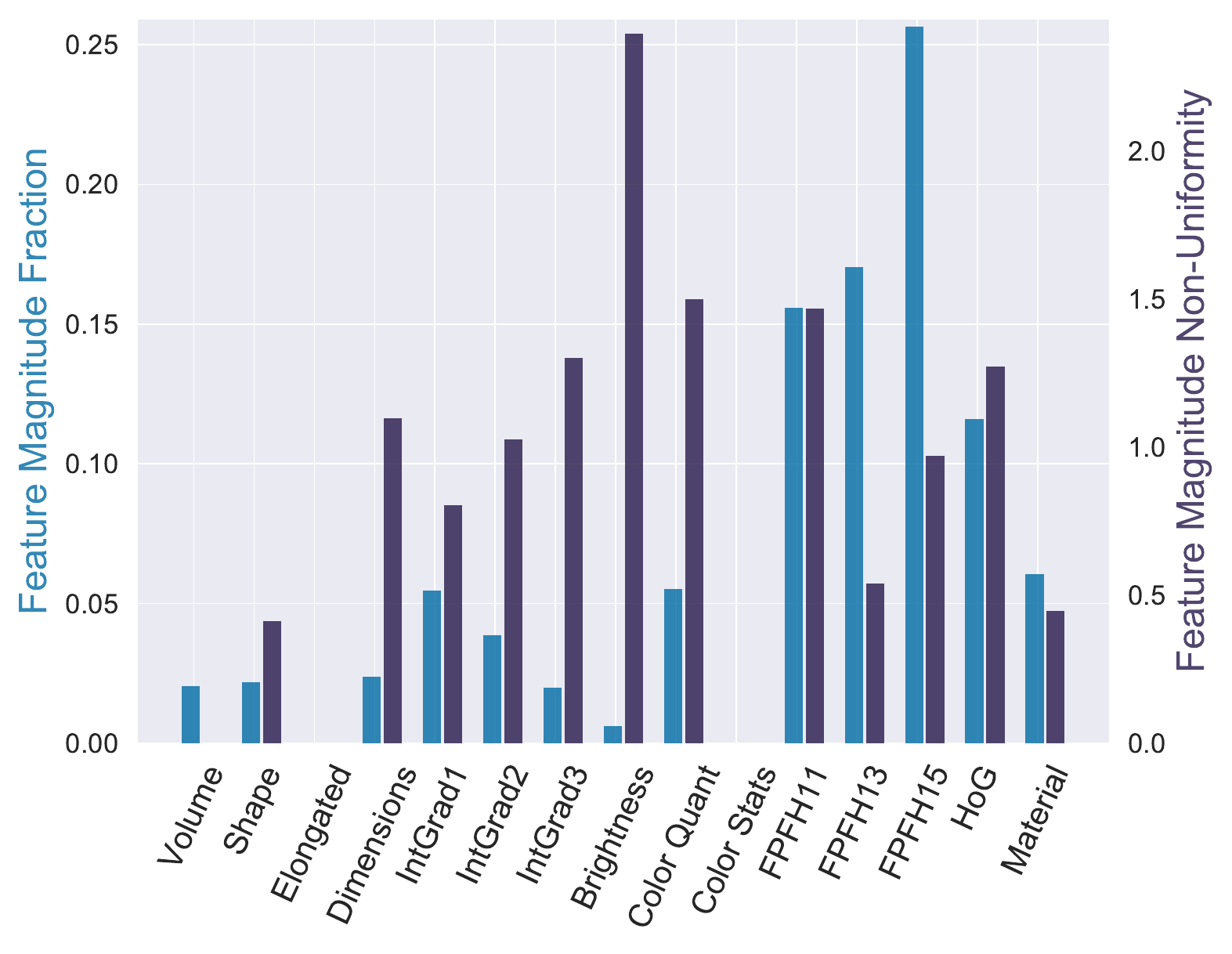}
        \caption{\footnotesize \textbf{Rolling} - The capability to roll is an intangible affordance dependent on the curvature of an object, however, not all objects that have curved surfaces affords rolling, for example, a cup with a handle. We expected the curvature features FPFH and HoG to be important which is the case. That FPFH15 (5cm search radius) is the most important feature is sensible since it describes curvature over big areas.}
        \label{fig::trRolling}
    \end{subfigure}
    \par\bigskip 
    \begin{subfigure}{0.48\textwidth}
        \includegraphics[width=\textwidth, clip, trim=7 8 7 6]{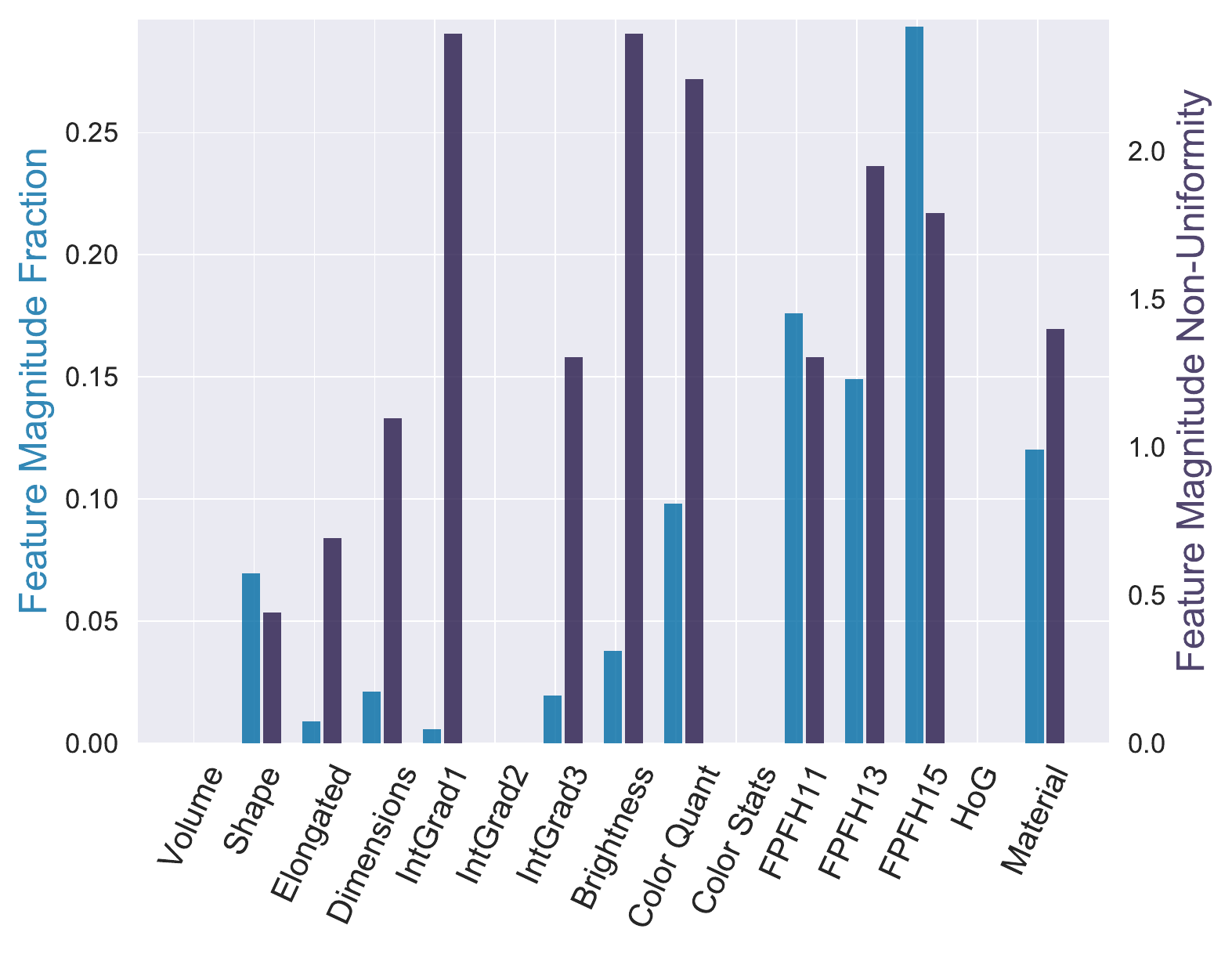}
        \caption{\footnotesize \textbf{Stacking} - Objects that afford stacking are typically cuboid shaped as indicated by the importance of the shape feature. Stackable objects also have flat surfaces. They should, therefore, like objects that afford rolling, be dependent on surface curvature features. A deeper analysis, in the projection section, instead shows that the edges of flat objects are the important factor. This is probably since many flats surfaces are not really flat due to noise in the depth camera.}
        \label{fig::trStacking}
    \end{subfigure}
    \caption{Barplots summarizing the importance of each feature for the affordances Rolling and Stacking. The left axis shows the sums of the magnitudes for each feature of the normalized transform, $\lVert\*L\rVert_{1}$. The right axis shows the KL-divergence between the normalized weights of a feature and a uniform distribution indicating the within feature distribution of magnitude values.}
    \label{fig::transformmags1}
\end{figure}

\begin{figure*}[!ht]
        \begin{subfigure}[t]{.49\textwidth}
        \includegraphics[width=\textwidth, clip, trim=7 8 7 6]{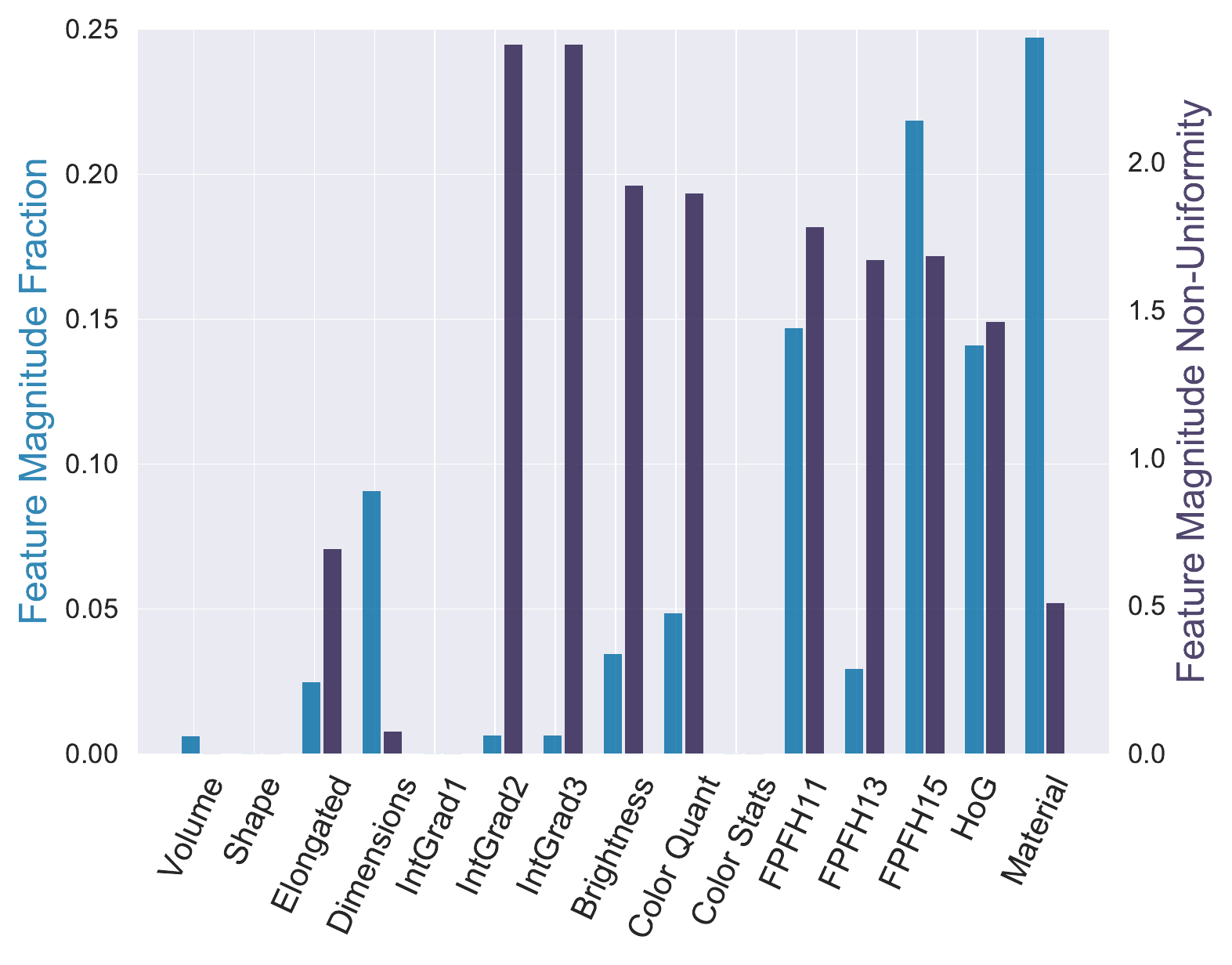}
        \caption{\footnotesize \textbf{Tools} - Tools are usually elongated and made of plastic, metal, or wood. They typically have handles which have a certain geometrical structure. All these aspects are reflected in the feature weights showing FPFH, HoG, and material as important features in addition to the dimension feature.}
        \label{fig::trTool}
    \end{subfigure}
    \hspace{0.2cm}
    \begin{subfigure}[t]{.49\textwidth}
        \includegraphics[width=\textwidth, clip, trim=7 8 7 6]{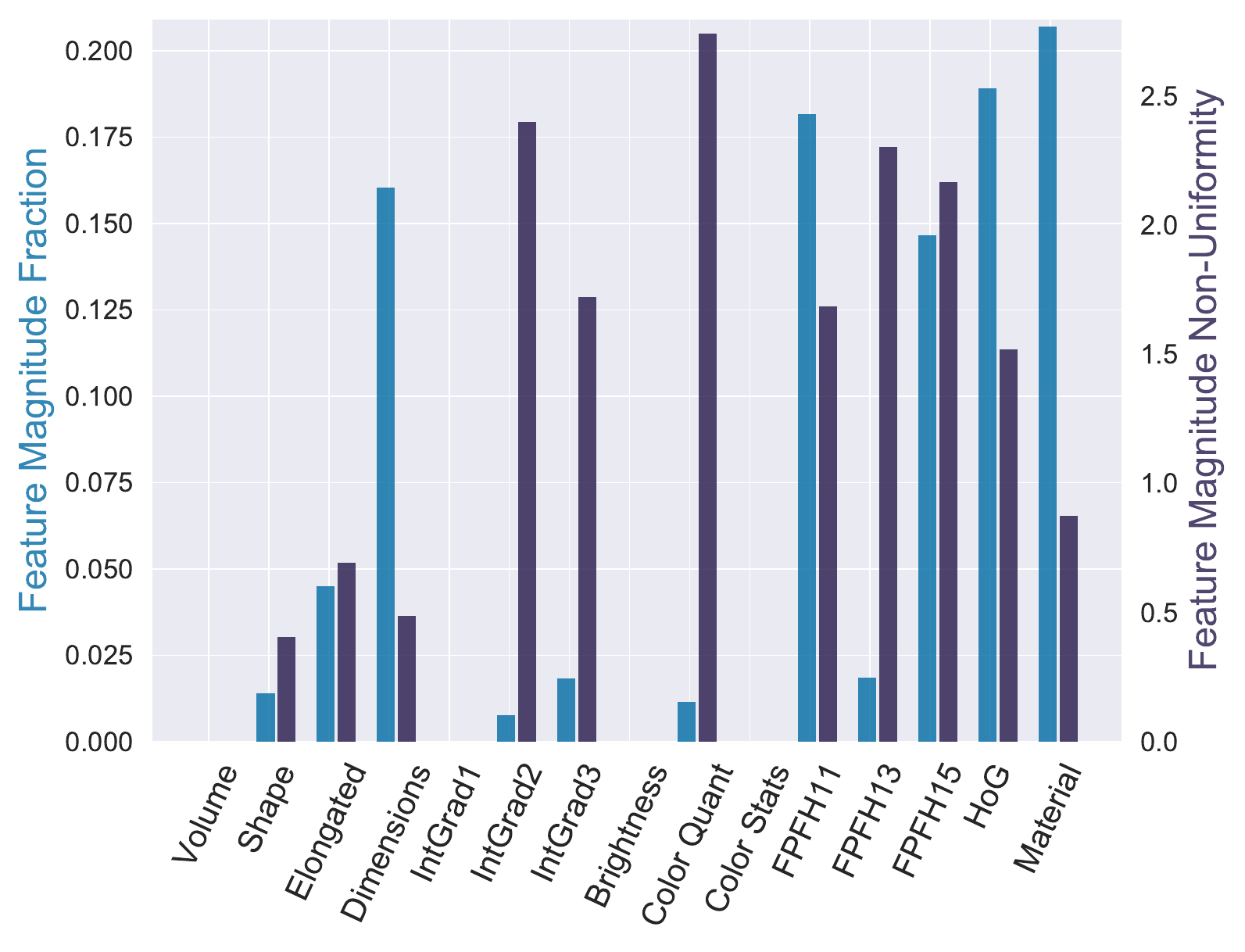}
        \caption{\footnotesize \textbf{Stirring} - Objects that affords stirring are elongated which means that the dimensions of an object and the elongation should be an important factor. The plot shows that dimensions are on par with the more complicated shape features.}
        \label{fig::trStirring}
    \end{subfigure}
    \begin{subfigure}[t]{.49\textwidth}
        \includegraphics[width=\textwidth, clip, trim=7 8 7 6]{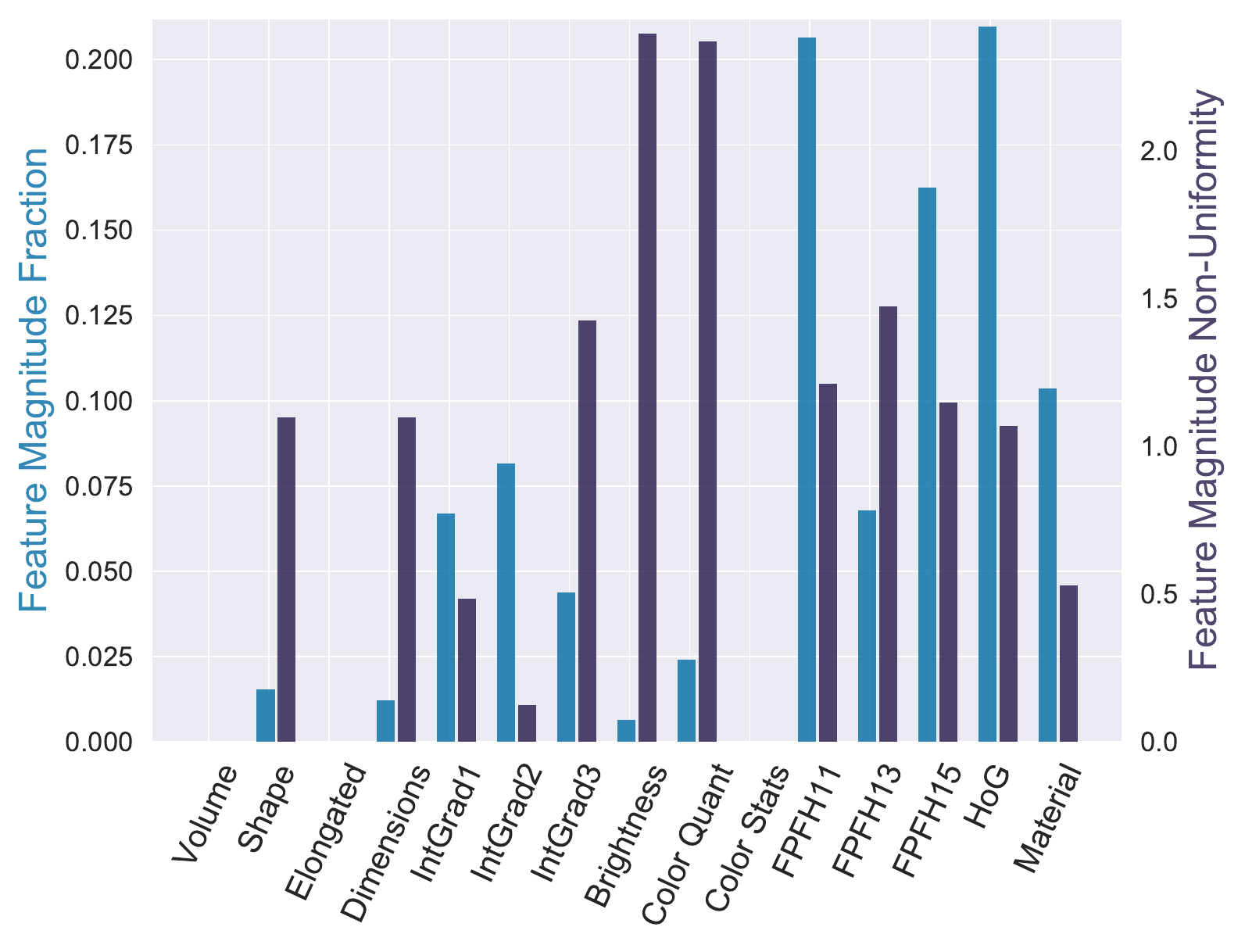}
        \caption{\footnotesize \textbf{Lifting Top} - Most objects with a top are cylindrical. This would explain the focus on the shape features FPFH and HoG. The gradient features are somewhat prominent indicating that objects with a top contain many lines, e.g. the clear line between top and body. Objects with a top usually have graphical labels that will give additional lines giving more weight to gradient features.}
        \label{fig::trLiftTop}
    \end{subfigure}
    \hspace{0.2cm}
    \begin{subfigure}[t]{.49\textwidth}
        \includegraphics[width=\textwidth, clip, trim=7 8 7 6]{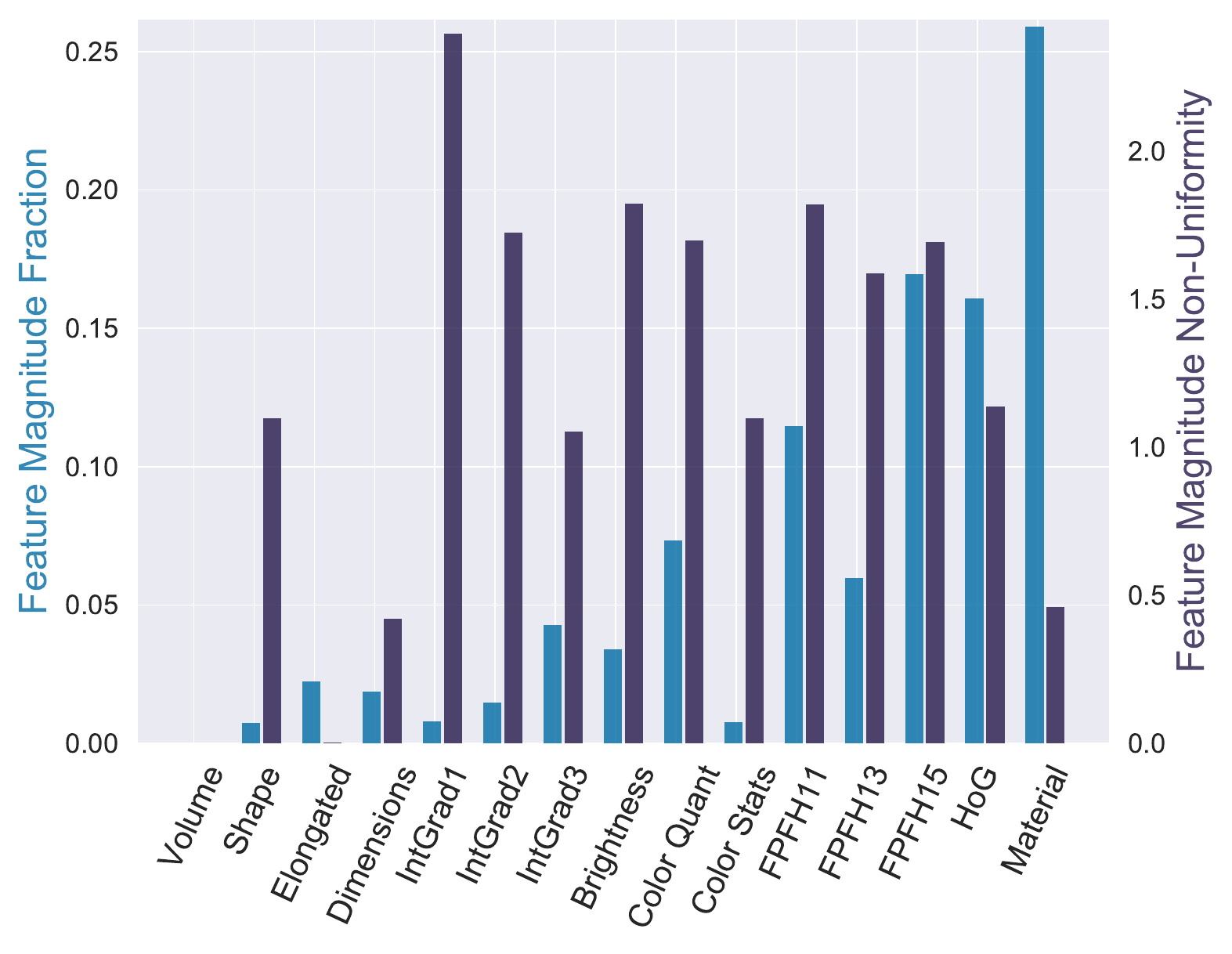}
        \caption{\footnotesize \textbf{Handle Grasping} - The histogram shows material features being the most important which is natural given that most objects are made of plastic, metal, or wood.  Given the excellent ability to locate handles by the model given in section \ref{section:feature_projection}  it seems the selected the parts of the FPFH features describes exactly the shape of handles on objects.}
        \label{fig::trHandleGrasp}
    \end{subfigure}
    \caption{Barplots summarizing the importance of each feature for the affordances Tools, Stirring, Lifting Top and Handle Grasping. The left axis shows the sums of the magnitudes for each feature of the normalized transform, $\lVert\*L\rVert_{1}$. The right axis shows the KL-divergence between the normalized weights of a feature and a uniform distribution indicating the within feature distribution of magnitude values.}
    \label{fig::transformmags2}
\end{figure*}

\subsection{Feature Selection}
We take the average of the $\*L$ column magnitudes over the $25$ runs and normalize, this will indicate each dimension's importance. We give results for $6$ of the more interesting affordances in Fig.\ref{fig::transformmags1}-\ref{fig::transformmags2}. The bar plots show the sum of the magnitude for each feature, that is, the fraction of each feature of the full magnitude vector. To provide a notion of the distribution of magnitude within each feature we compute the KL-divergence of the normalized magnitudes for the features with a uniform distribution. The right-hand bars thus indicate how evenly the magnitudes are distributed across each of the features.

The general tendency is that some features like material, size, and shape are important across the board. Size and material are good for making an initial guess. For example, there are no tools made of paper or very thin objects that affords Stacking or Handle Grasping, etc. Shape features are more specific and vary much more across the different affordances, however, in general size, shape, and material features are the most important as expected. Analyzing the diagrams for all the affordances it is clear that the features *volume, shape primitive, gradients, and color stats* are not as important for classifying affordances compared to the other features. 

\renewcommand{\thesubfigure}{\thefigure.\arabic{subfigure}}
\begin{figure*}
\setlength\tabcolsep{0.5pt}

\begin{tabular}{cccc}
    \begin{subfigure}{0.2465\textwidth}
        \includegraphics[width=0.98\textwidth,clip,trim=0 0 0 0, cfbox=Black 0.5pt 0pt]{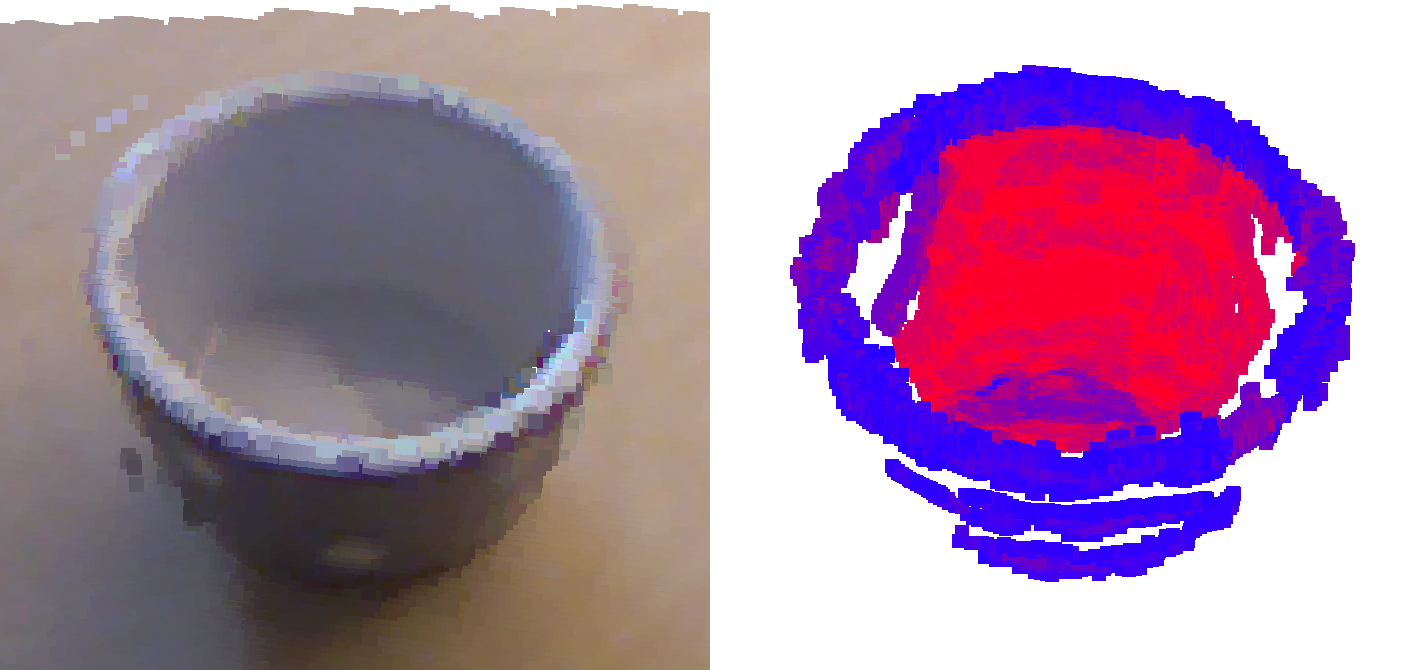}
        \caption{Drinking}\label{fig:feature_projections_drinking1}
    \end{subfigure} &
    \begin{subfigure}{0.2465\textwidth}
        \includegraphics[width=0.98\textwidth,clip,trim=0 0 0 0, cfbox=Black 0.5pt 0pt]{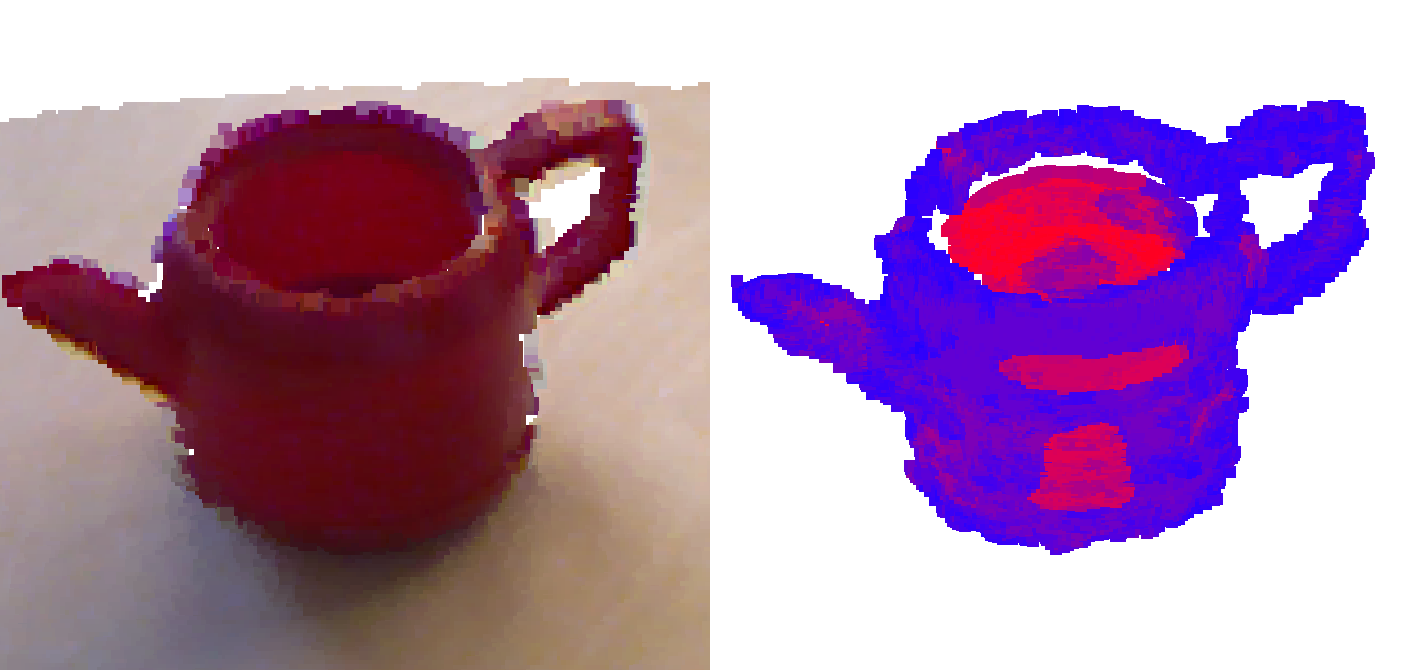}
        \caption{Drinking}\label{fig:feature_projections_drinking2}
    \end{subfigure} &
    \begin{subfigure}{0.2465\textwidth}
        \includegraphics[width=0.98\textwidth,clip,trim=0 0 0 0, cfbox=Black 0.5pt 0pt]{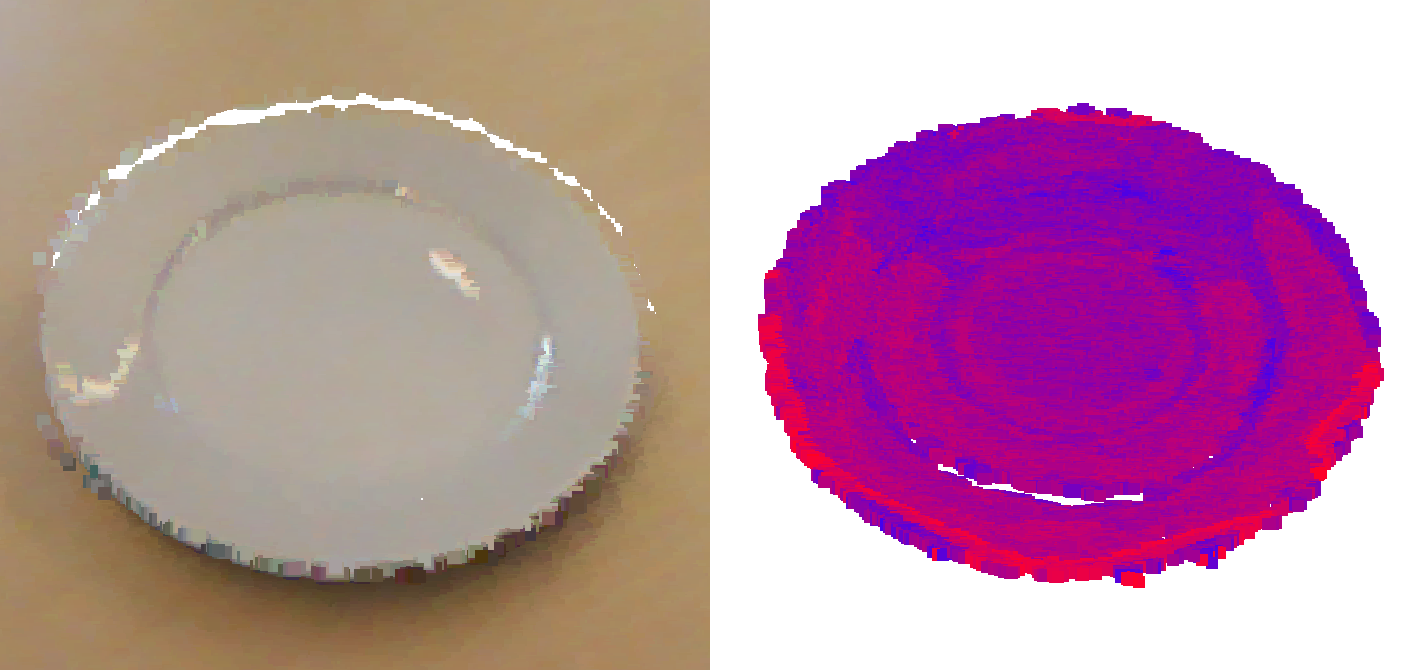}
        \caption{Eating From}\label{fig:feature_projections_eatingfrom1}
    \end{subfigure} &
    \begin{subfigure}{0.2465\textwidth}
        \includegraphics[width=0.98\textwidth,clip,trim=0 0 0 0, cfbox=Black 0.5pt 0pt]{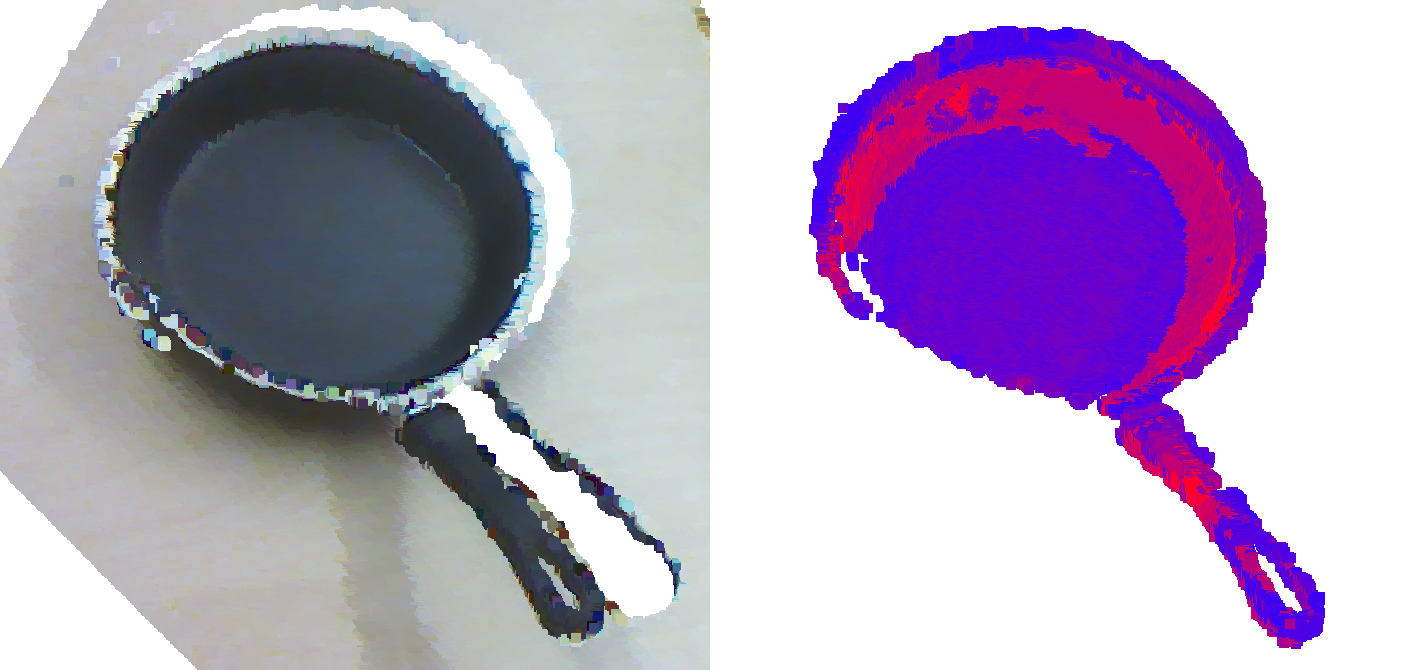}
        \caption{Eating From}\label{fig:feature_projections_eatingfrom2}
    \end{subfigure} \\[8.ex]
    \begin{subfigure}{0.2465\textwidth}
        \includegraphics[width=0.98\textwidth,clip,trim=0 0 0 0, cfbox=Black 0.5pt 0pt]{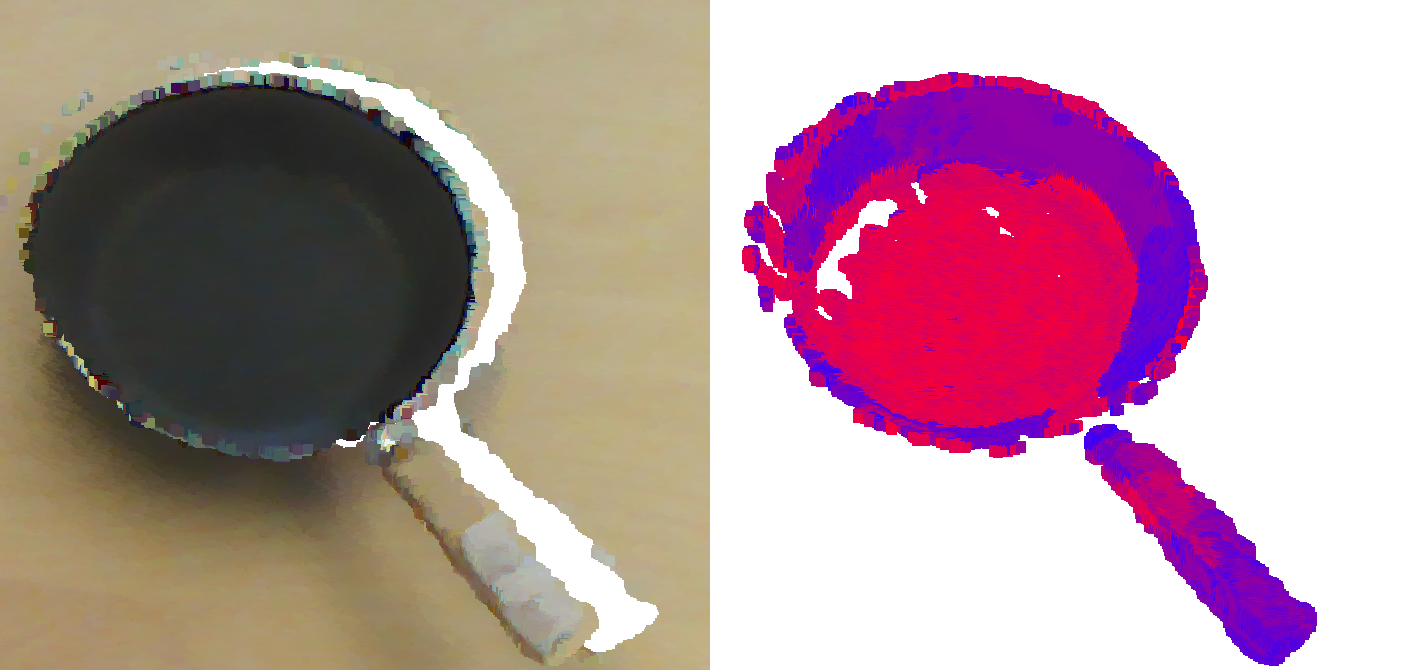}
        \caption{Eating From}\label{fig:feature_projections_eatingfrom3}
    \end{subfigure} &
    \begin{subfigure}{0.2465\textwidth}
        \includegraphics[width=0.98\textwidth,clip,trim=0 0 0 0, cfbox=Black 0.5pt 0pt]{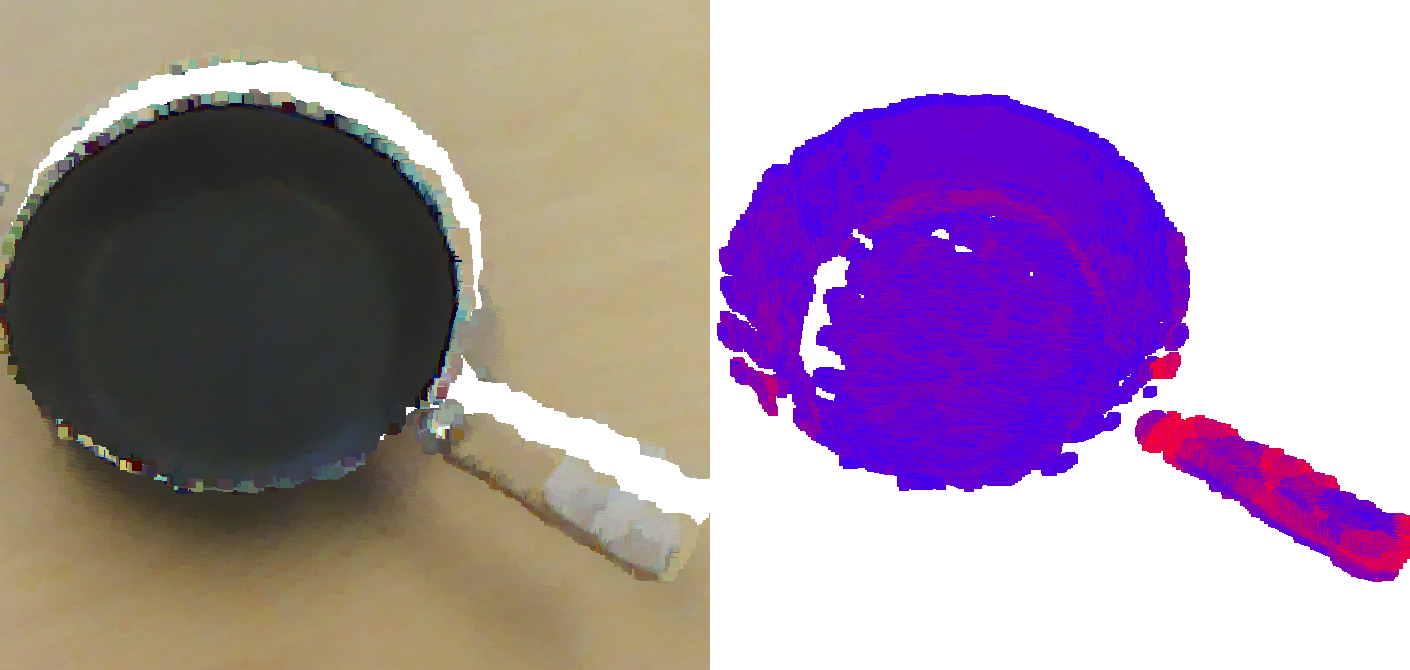}
        \caption{Handle Grasping}\label{fig:feature_projections_handlegrasping1}
    \end{subfigure} &
    \begin{subfigure}{0.2465\textwidth}
        \includegraphics[width=0.98\textwidth,clip,trim=0 0 0 0, cfbox=Black 0.5pt 0pt]{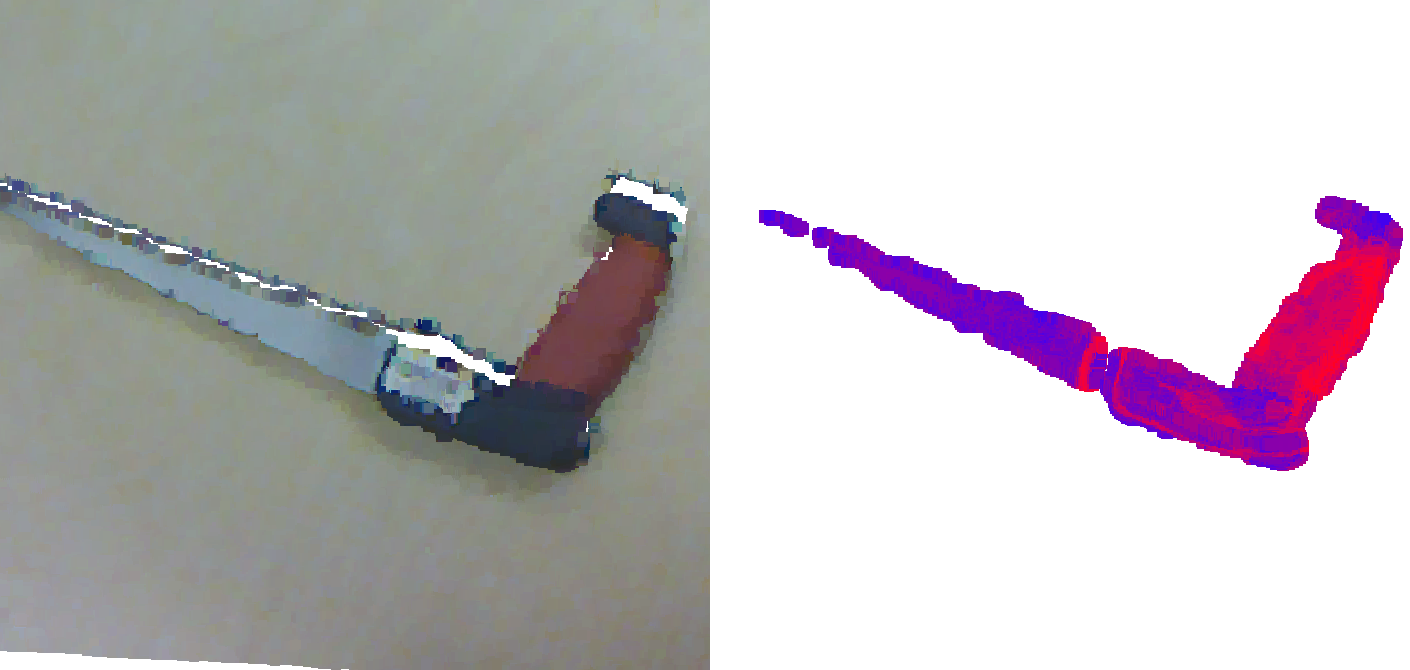}
        \caption{Handle Grasping}\label{fig:feature_projections_handlegrasping2}
    \end{subfigure} &
    \begin{subfigure}{0.2465\textwidth}
        \includegraphics[width=0.98\textwidth,clip,trim=0 0 0 0, cfbox=Black 0.5pt 0pt]{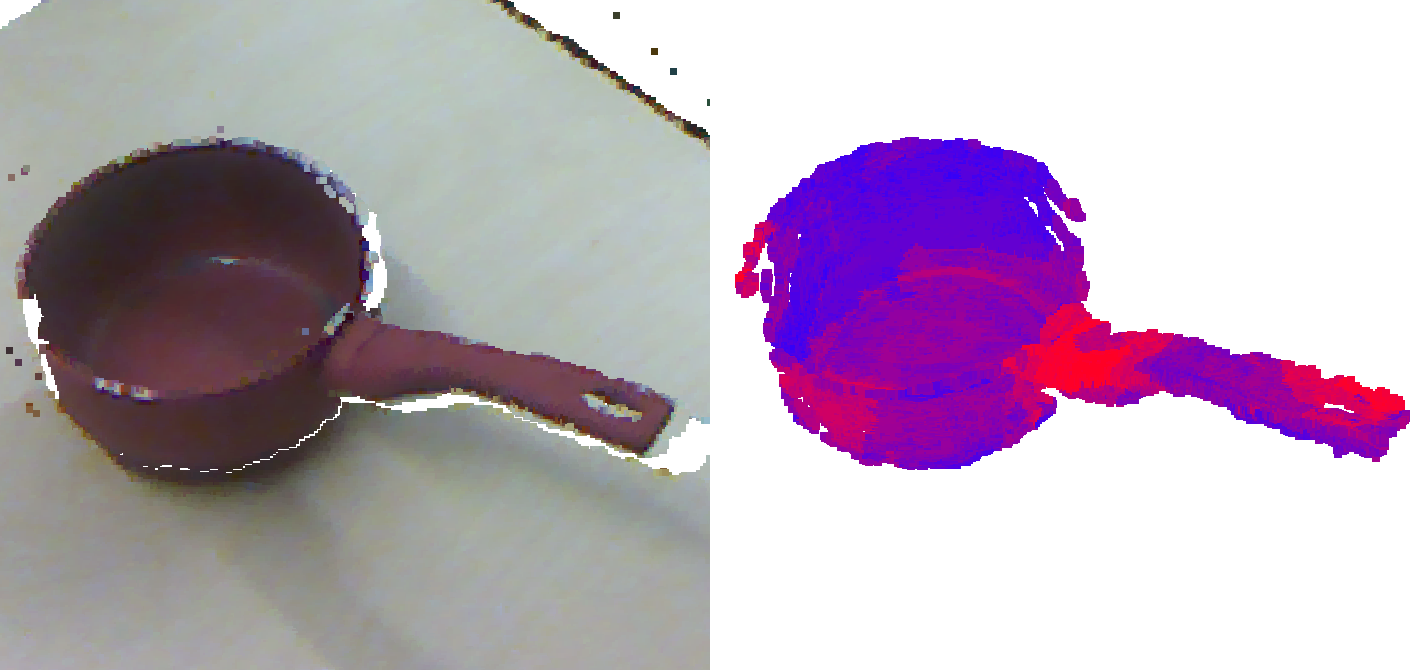}
        \caption{Handle Grasping}\label{fig:feature_projections_handlegrasping3}
    \end{subfigure} \\[8.ex]
    \begin{subfigure}{0.2465\textwidth}
        \includegraphics[width=0.98\textwidth,clip,trim=0 0 0 0, cfbox=Black 0.5pt 0pt]{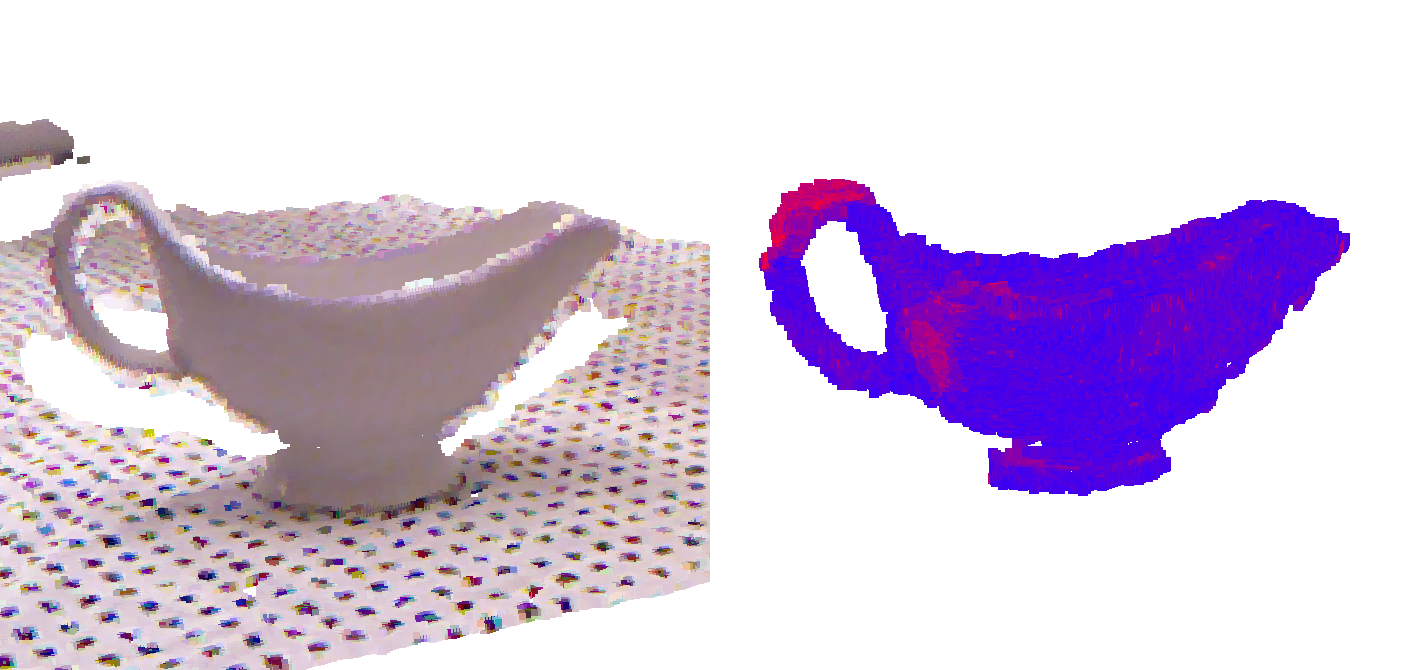}
        \caption{Hanging}\label{fig:feature_projections_hanging1}
    \end{subfigure} &
    \begin{subfigure}{0.2465\textwidth}
        \includegraphics[width=0.98\textwidth,clip,trim=0 0 0 0, cfbox=Black 0.5pt 0pt]{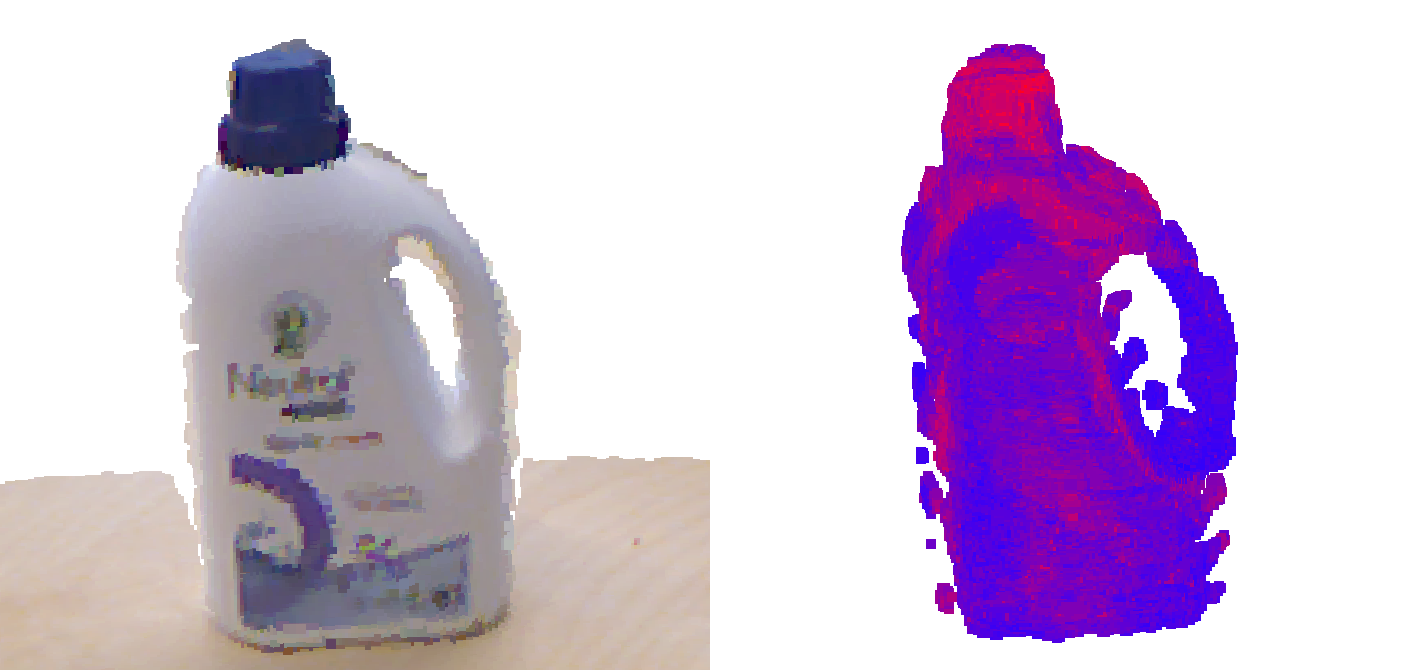}
        \caption{Hanging}\label{fig:feature_projections_hanging2}
    \end{subfigure} &
    \begin{subfigure}{0.2465\textwidth}
        \includegraphics[width=0.98\textwidth,clip,trim=0 0 0 0, cfbox=Black 0.5pt 0pt]{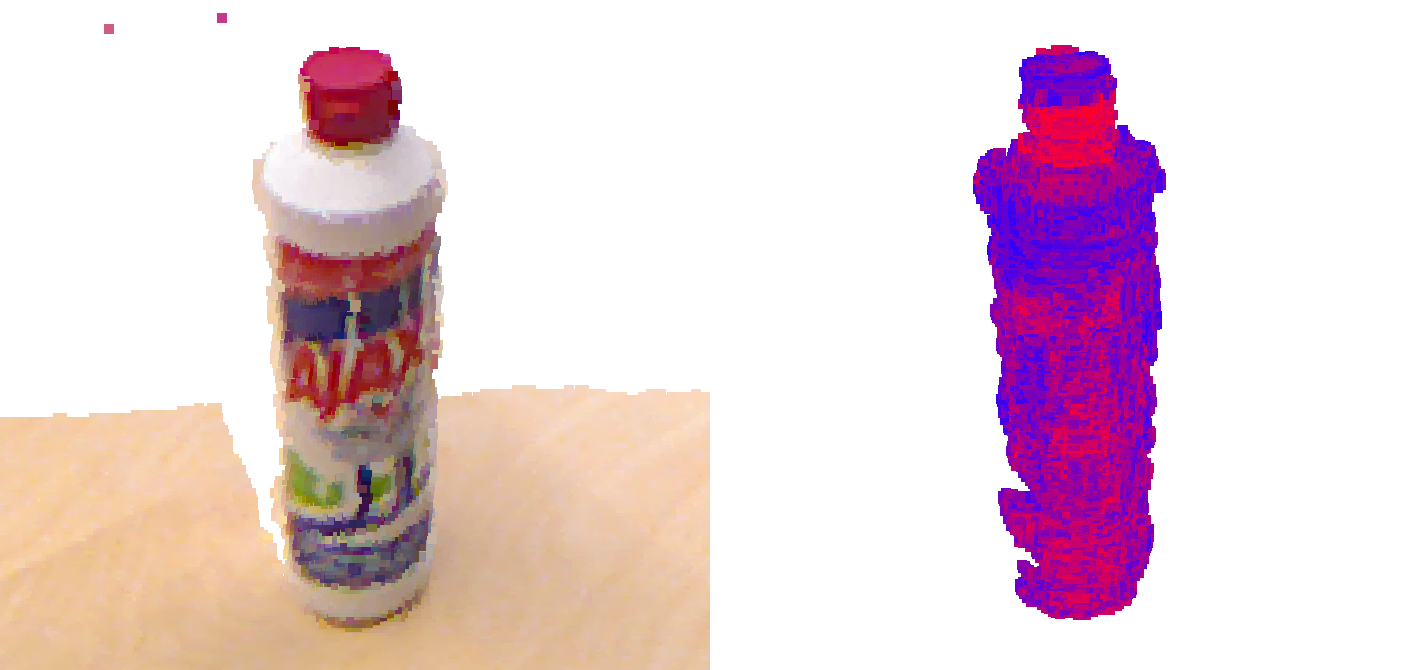}
        \caption{Lifting Top}\label{fig:feature_projections_liftingtop1}
    \end{subfigure} &
    \begin{subfigure}{0.2465\textwidth}
        \includegraphics[width=0.98\textwidth,clip,trim=0 0 0 0, cfbox=Black 0.5pt 0pt]{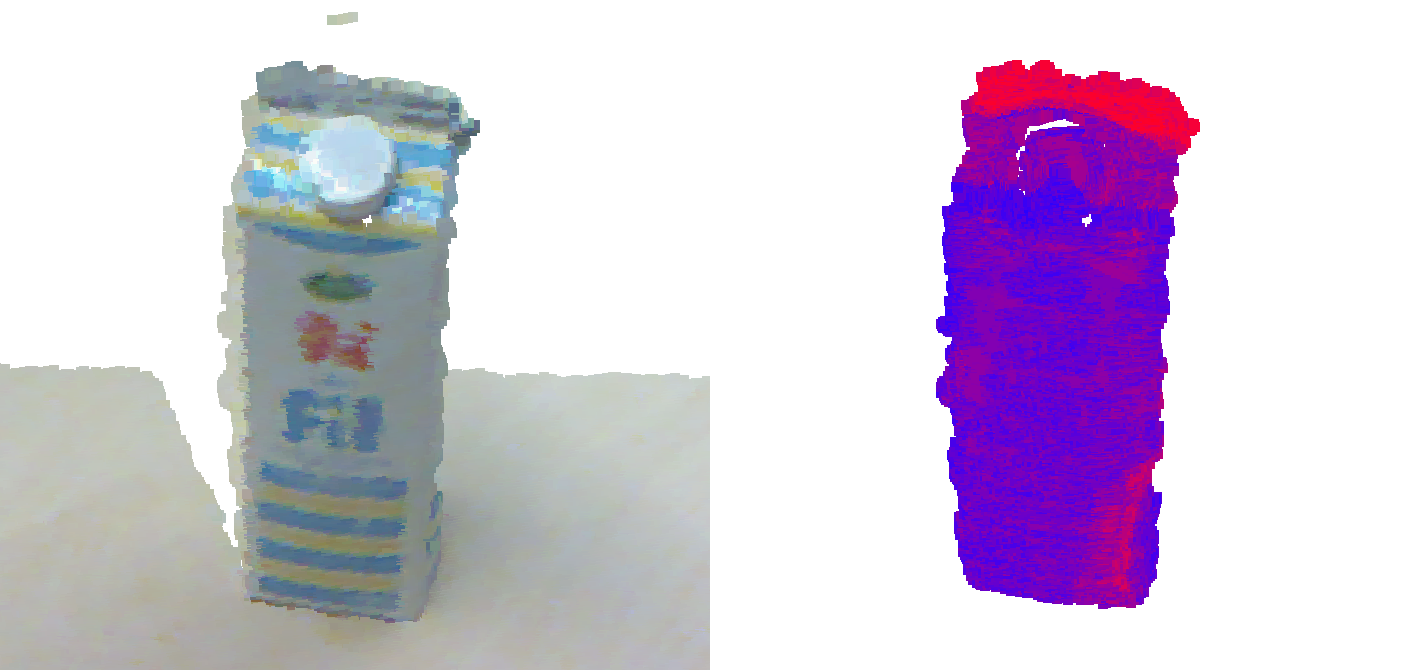}
        \caption{Lifting Top}\label{fig:feature_projections_liftingtop2}
    \end{subfigure} \\[8.ex]
    \begin{subfigure}{0.2465\textwidth}
        \includegraphics[width=0.98\textwidth,clip,trim=0 0 0 0, cfbox=Black 0.5pt 0pt]{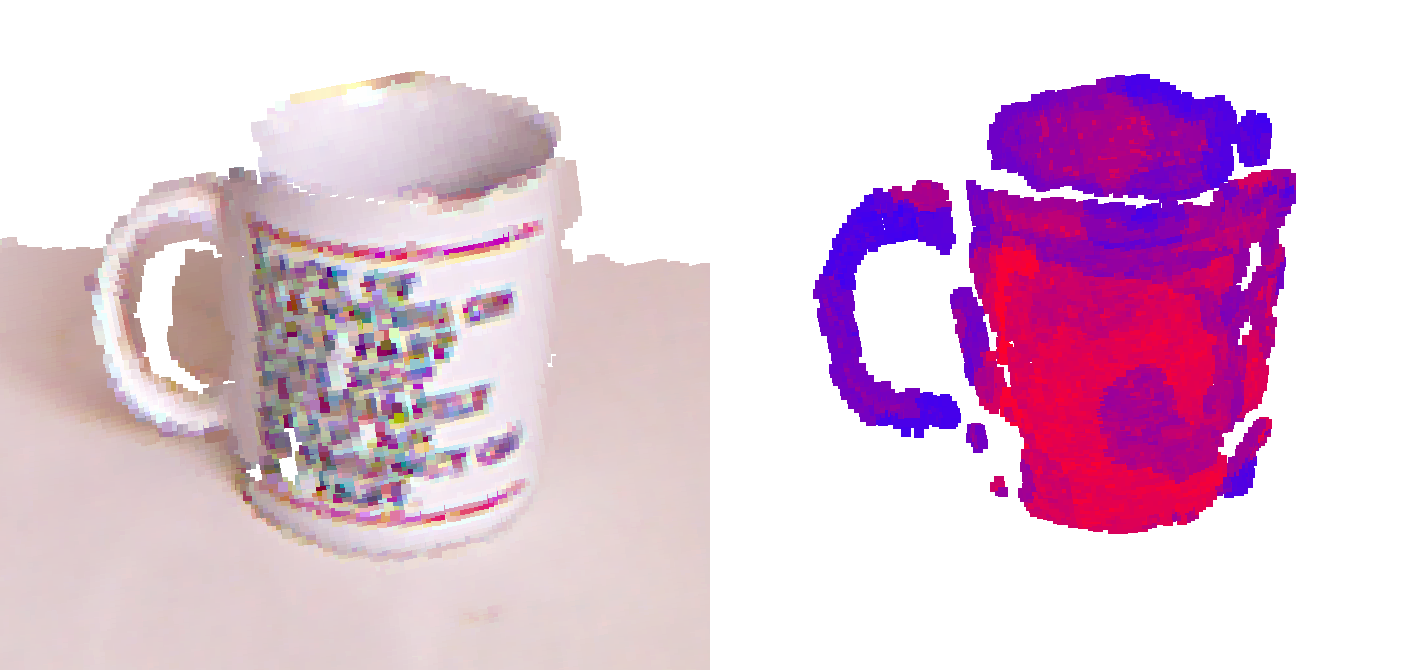}
        \caption{Loop Grasping}\label{fig:feature_projections_loopgrasping1}
    \end{subfigure} &
    \begin{subfigure}{0.2465\textwidth}
        \includegraphics[width=0.98\textwidth,clip,trim=0 0 0 0, cfbox=Black 0.5pt 0pt]{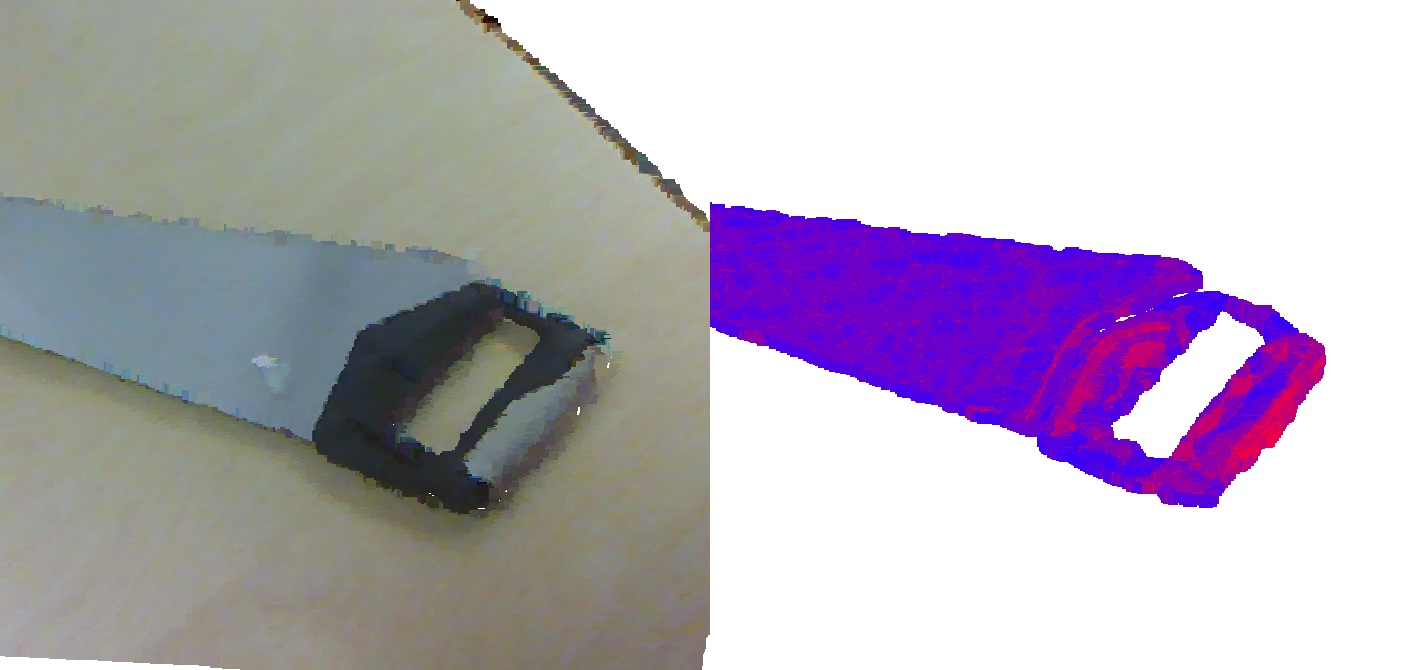}
        \caption{Loop Grasping}\label{fig:feature_projections_loopgrasping2}
    \end{subfigure} &
    \begin{subfigure}{0.2465\textwidth}
        \includegraphics[width=0.98\textwidth,clip,trim=0 0 0 0, cfbox=Black 0.5pt 0pt]{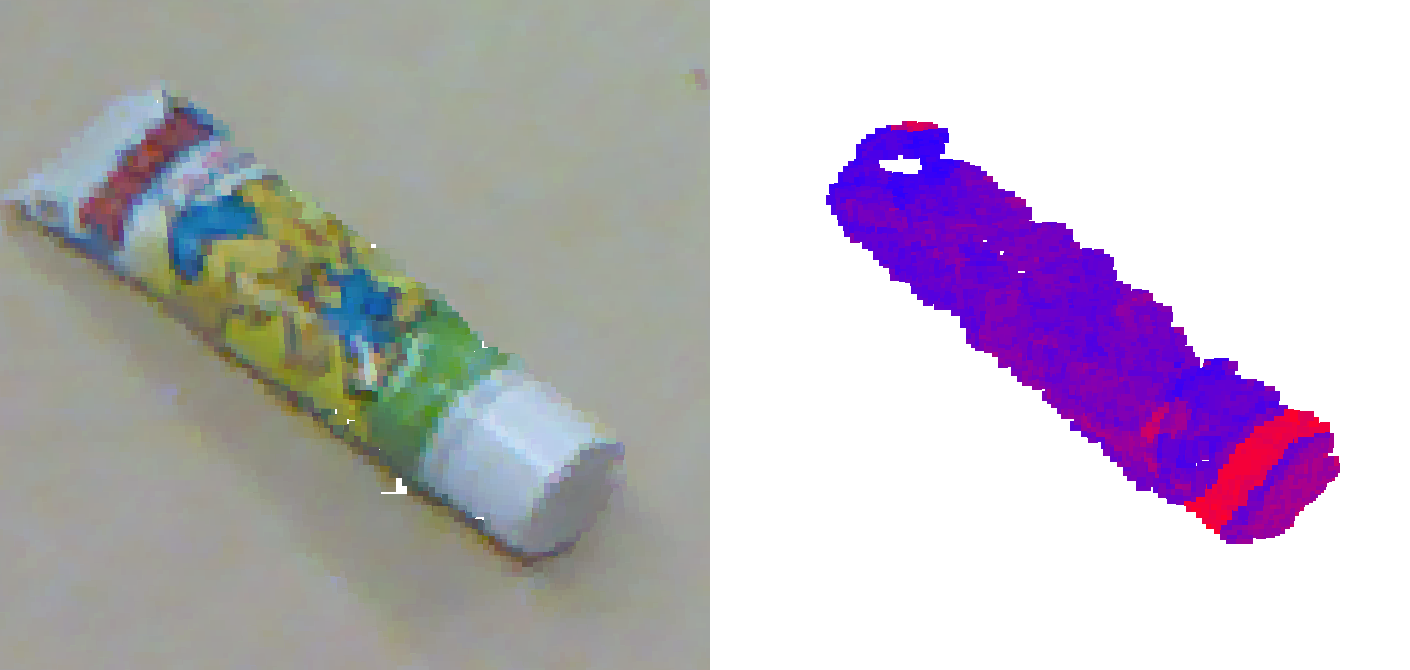}
        \caption{Opening}\label{fig:feature_projections_opening1}
    \end{subfigure} &
    \begin{subfigure}{0.2465\textwidth}
        \includegraphics[width=0.98\textwidth,clip,trim=0 0 0 0, cfbox=Black 0.5pt 0pt]{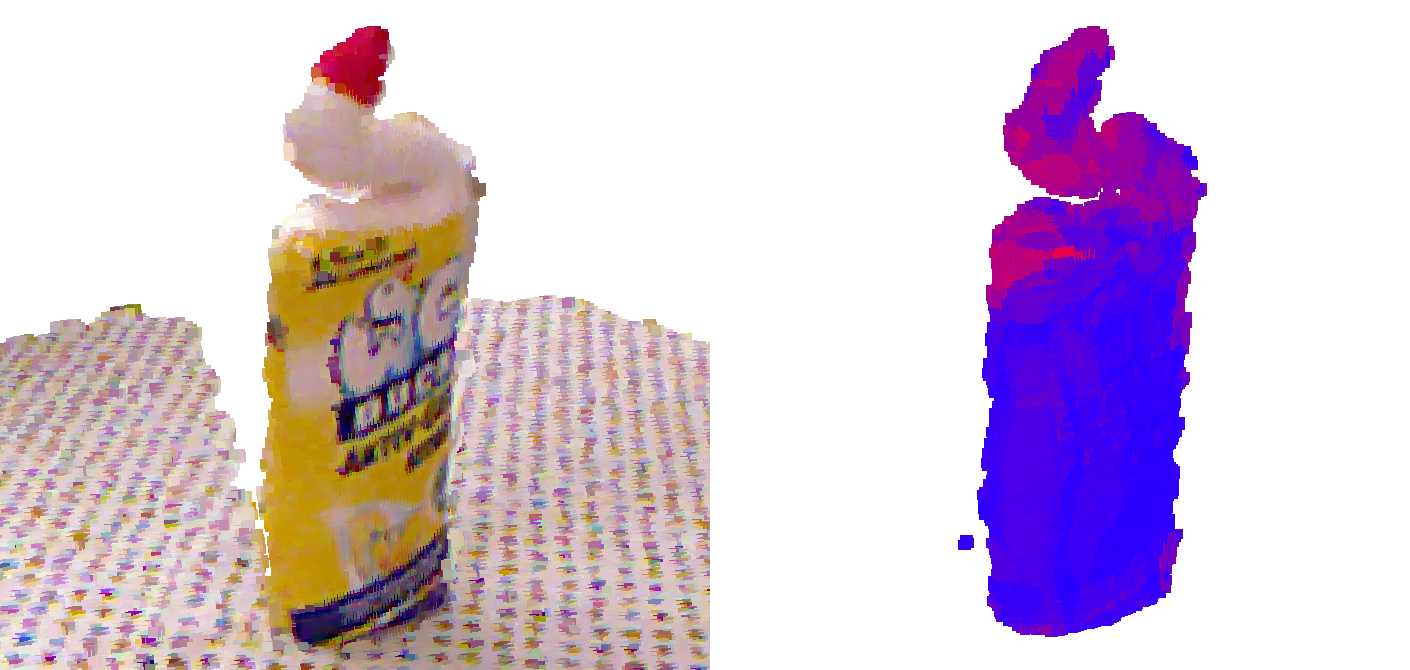}
        \caption{Opening}\label{fig:feature_projections_opening2}
    \end{subfigure} \\[8.ex]
    \begin{subfigure}{0.2465\textwidth}
        \includegraphics[width=0.98\textwidth,clip,trim=0 0 0 0, cfbox=Black 0.5pt 0pt]{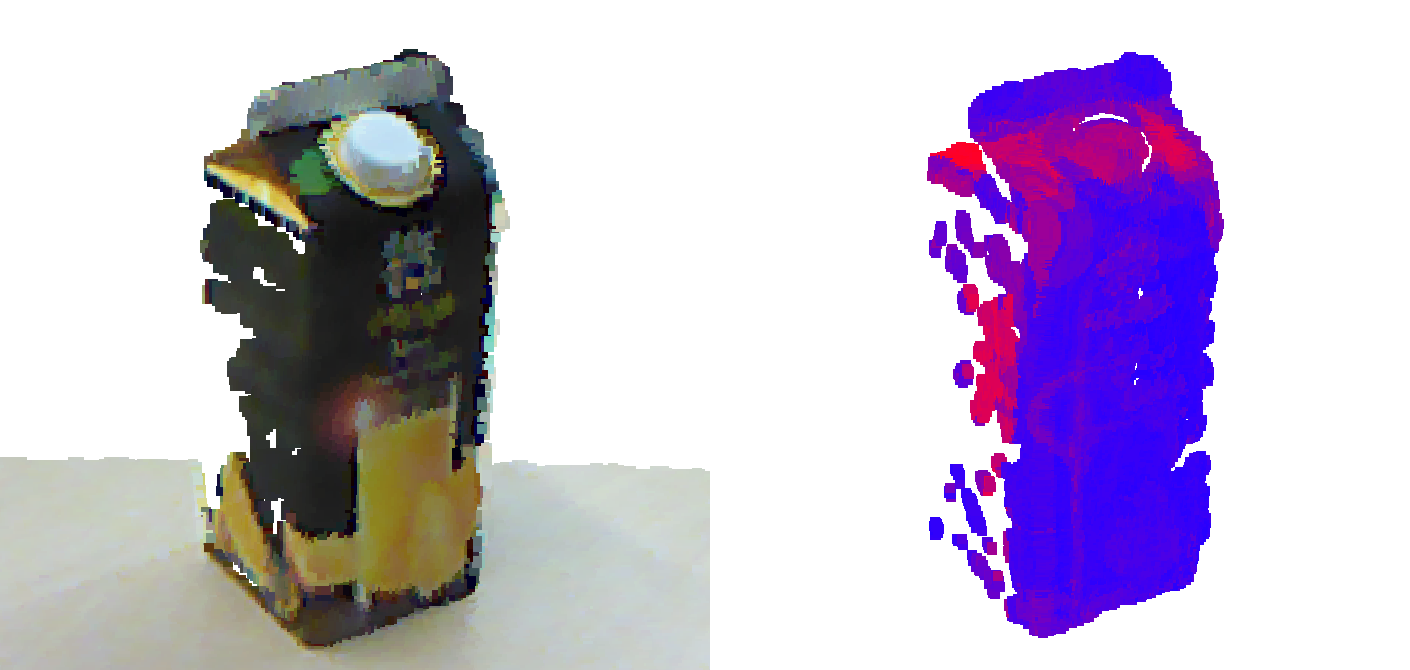}
        \caption{Opening}\label{fig:feature_projections_opening3}
    \end{subfigure} &
    \begin{subfigure}{0.2465\textwidth}
        \includegraphics[width=0.98\textwidth,clip,trim=0 0 0 0, cfbox=Black 0.5pt 0pt]{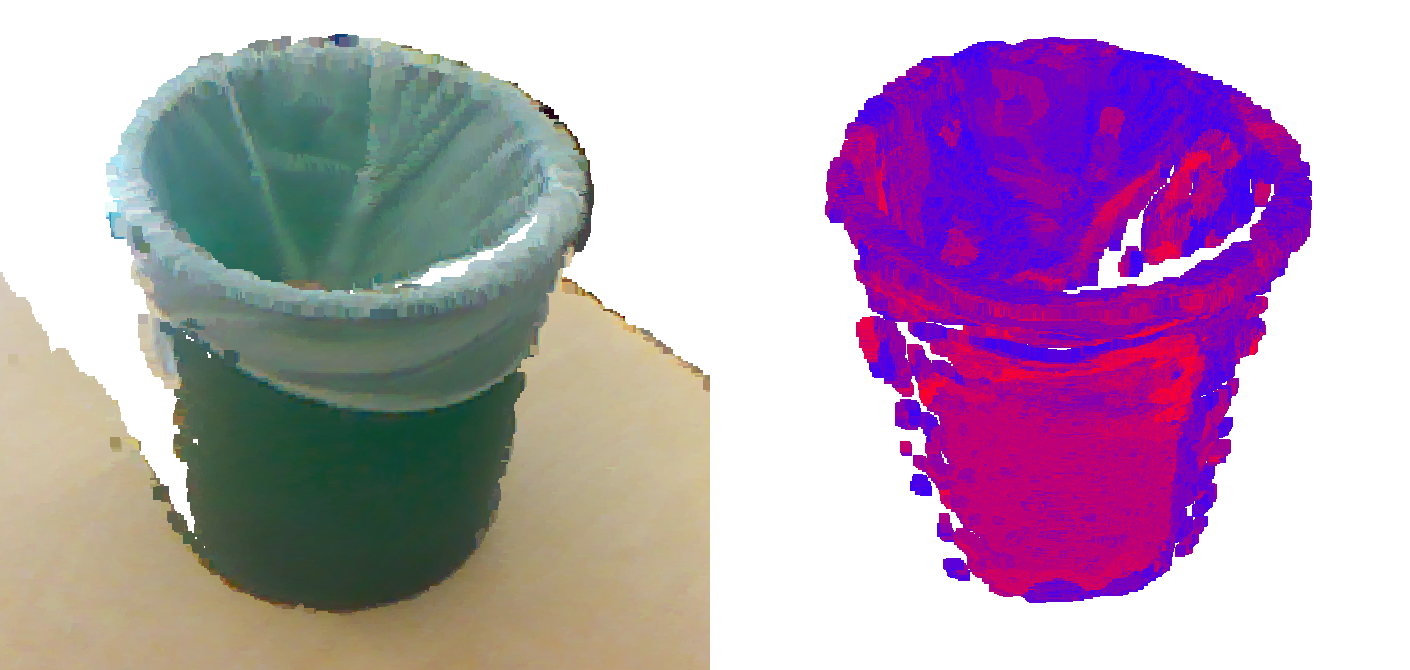}
        \caption{Putting}\label{fig:feature_projections_putting1}
    \end{subfigure} &
    \begin{subfigure}{0.2465\textwidth}
        \includegraphics[width=0.98\textwidth,clip,trim=0 0 0 0, cfbox=Black 0.5pt 0pt]{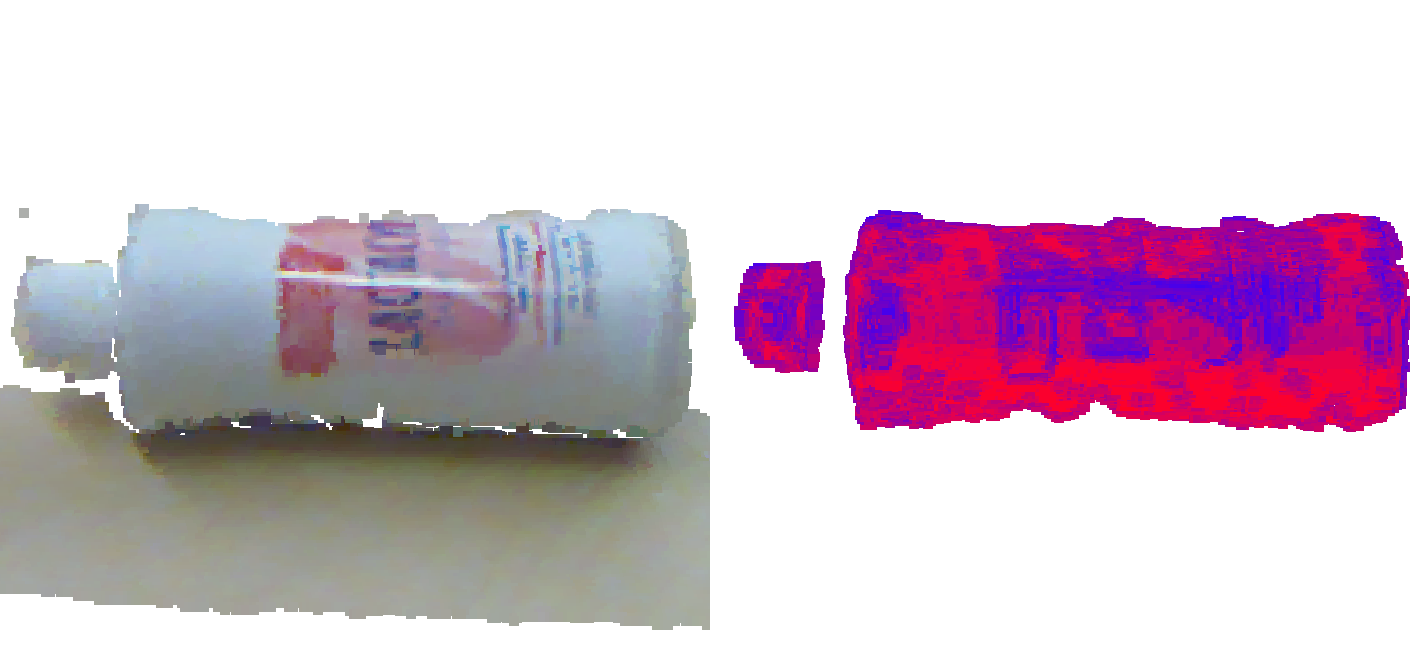}
        \caption{Rolling}\label{fig:feature_projections_rolling1}
    \end{subfigure} &
    \begin{subfigure}{0.2465\textwidth}
        \includegraphics[width=0.98\textwidth,clip,trim=0 0 0 0, cfbox=Black 0.5pt 0pt]{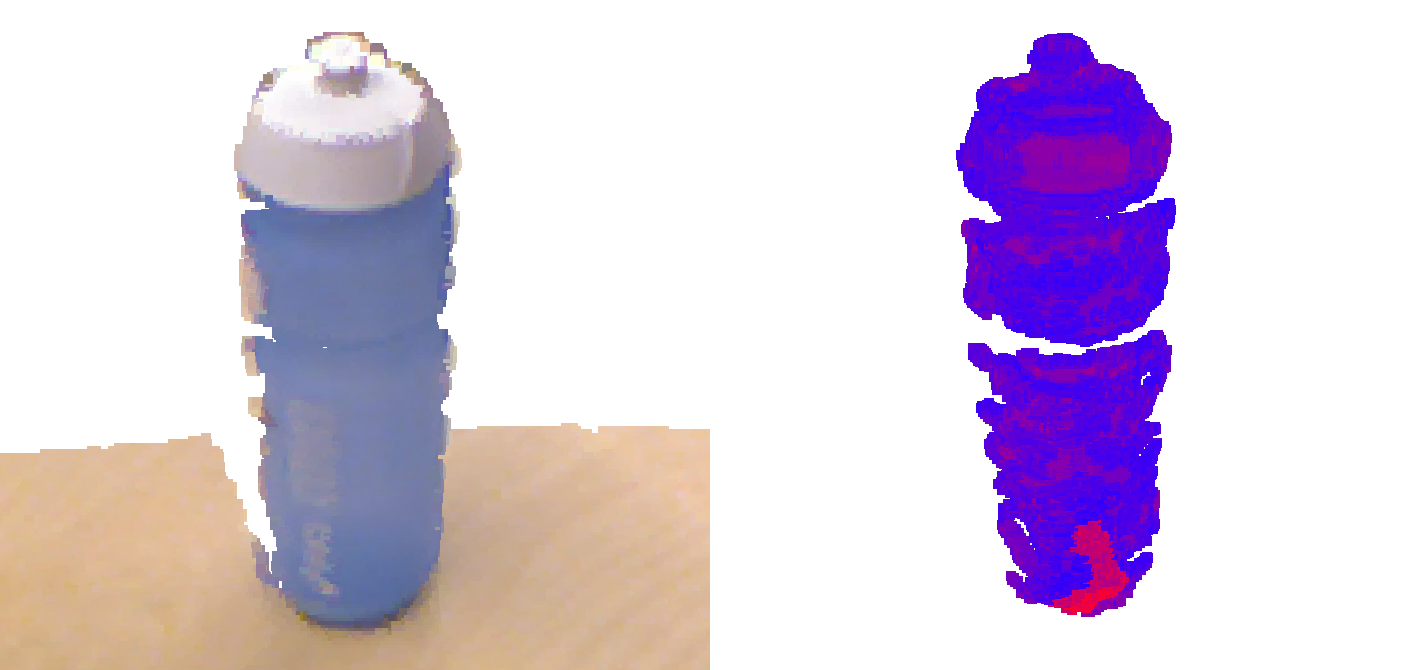}
        \caption{Rolling}\label{fig:feature_projections_rolling2}
    \end{subfigure} \\[8.ex]
    \begin{subfigure}{0.2465\textwidth}
        \includegraphics[width=0.98\textwidth,clip,trim=0 0 0 0, cfbox=Black 0.5pt 0pt]{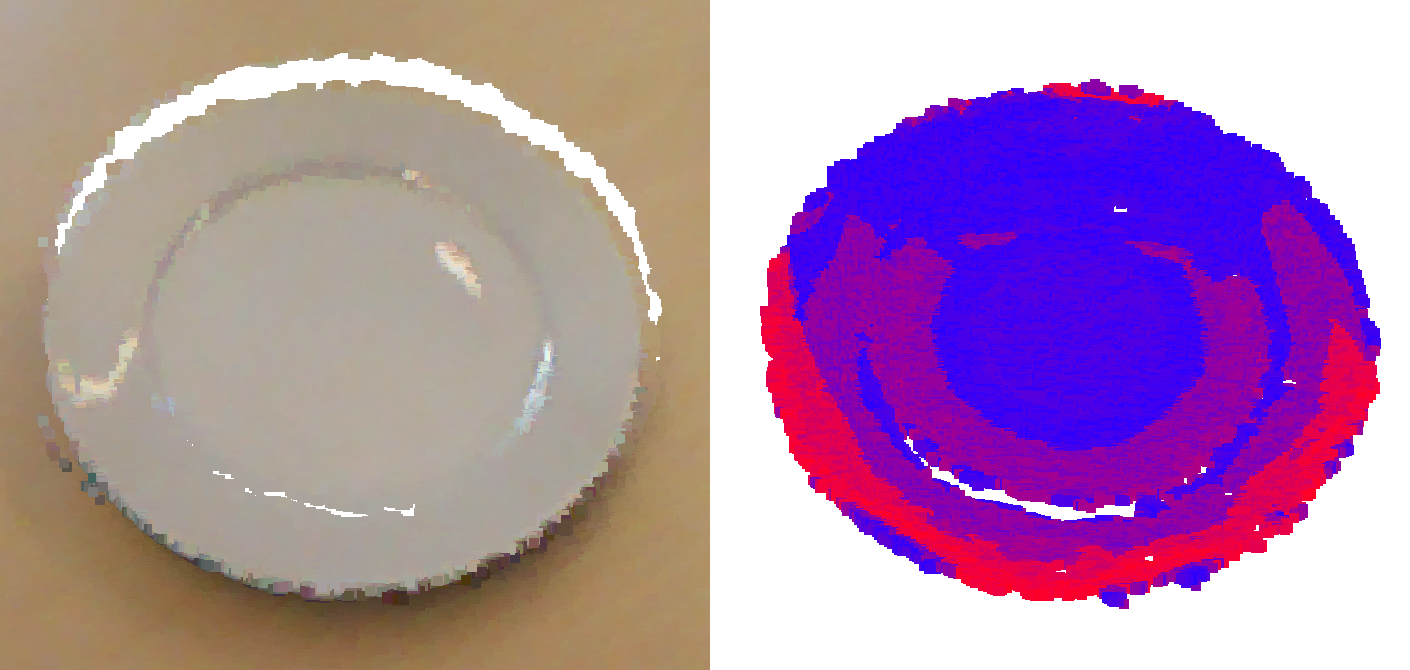}
        \caption{Stacking}\label{fig:feature_projections_stacking1}
    \end{subfigure} &
    \begin{subfigure}{0.2465\textwidth}
        \includegraphics[width=0.98\textwidth,clip,trim=0 0 0 0, cfbox=Black 0.5pt 0pt]{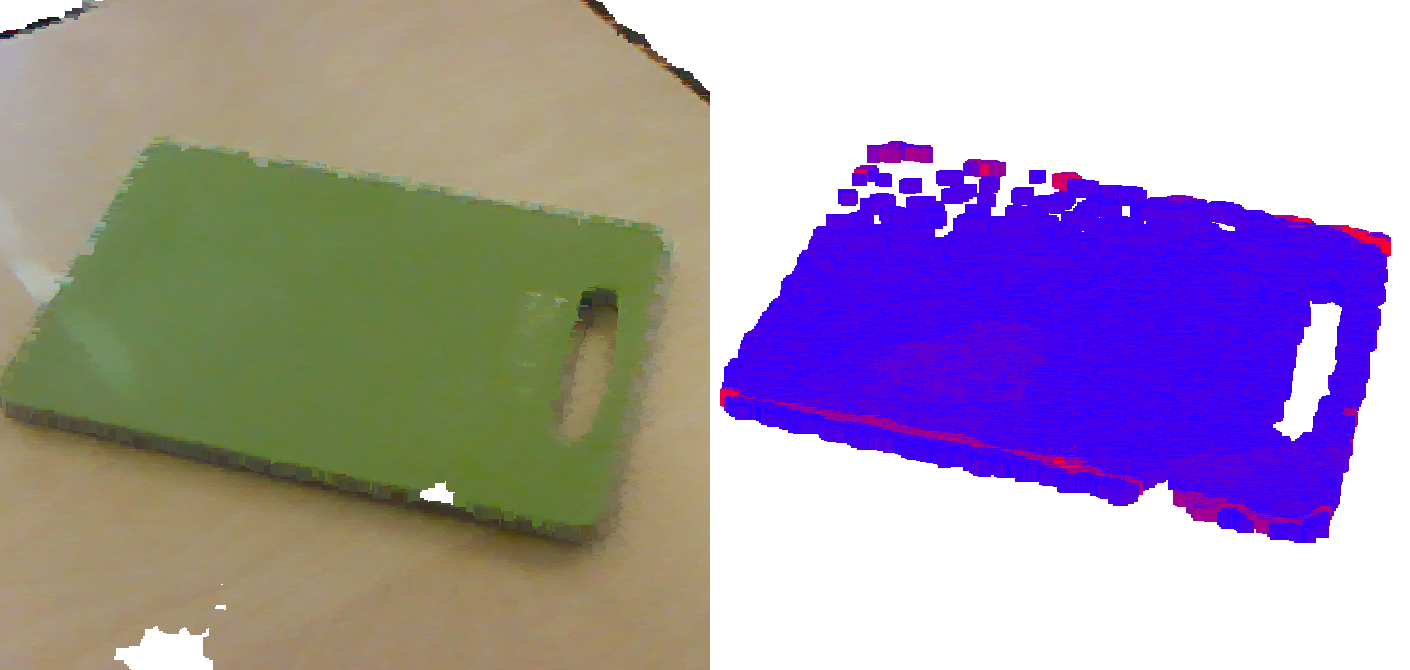}
        \caption{Stacking}\label{fig:feature_projections_stacking2}
    \end{subfigure} &
    \begin{subfigure}{0.2465\textwidth}
        \includegraphics[width=0.98\textwidth,clip,trim=0 0 0 0, cfbox=Black 0.5pt 0pt]{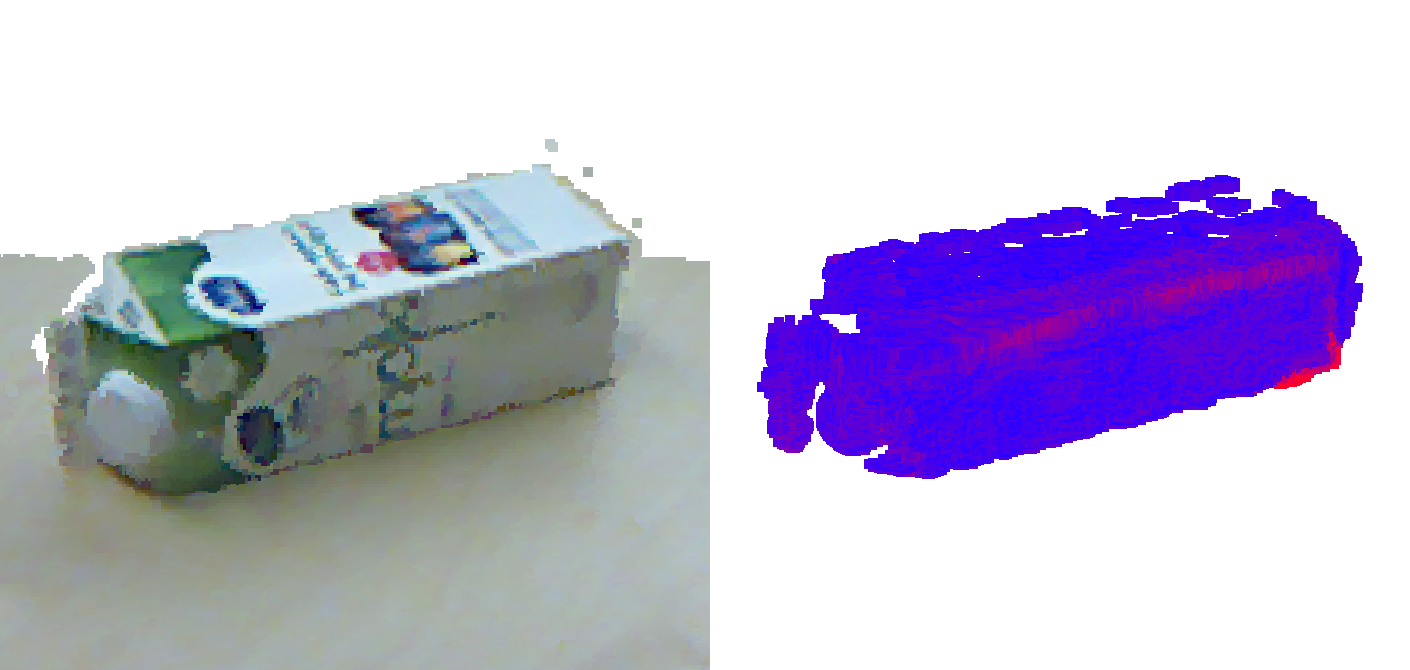}
        \caption{Stacking}\label{fig:feature_projections_stacking3}
    \end{subfigure} &
    \begin{subfigure}{0.2465\textwidth}
        \includegraphics[width=0.98\textwidth,clip,trim=0 0 0 0, cfbox=Black 0.5pt 0pt]{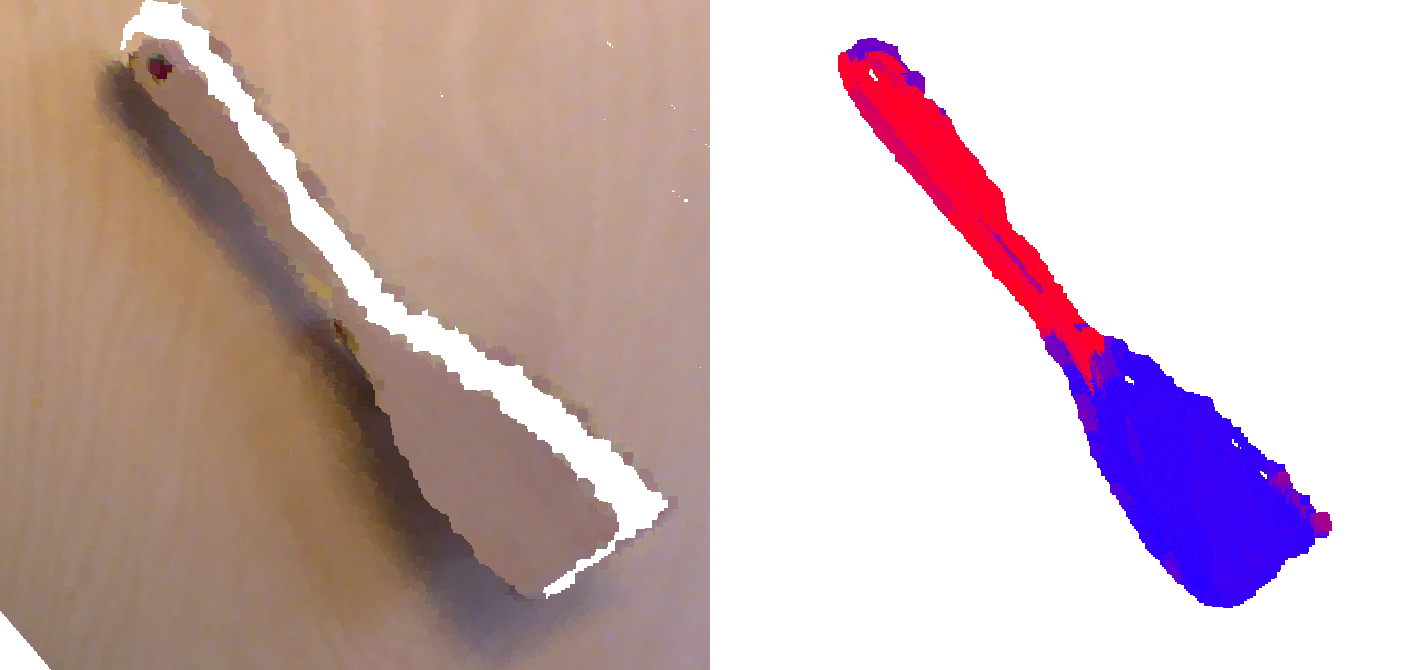}
        \caption{Stirring}\label{fig:feature_projections_stirring1}
    \end{subfigure}  \\[8.ex]
    \begin{subfigure}{0.2465\textwidth}
        \includegraphics[width=0.98\textwidth,clip,trim=0 0 0 0, cfbox=Black 0.5pt 0pt]{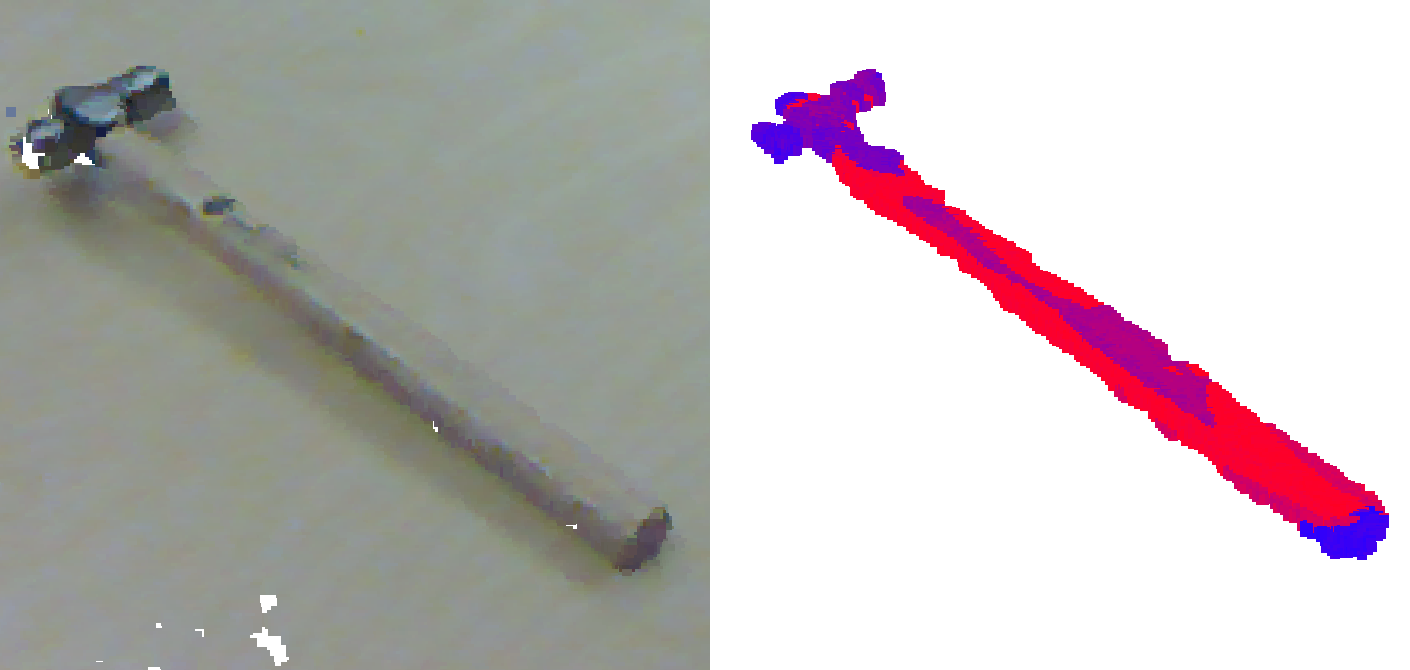}
        \caption{Stirring}\label{fig:feature_projections_stirring2}
    \end{subfigure} &
    \begin{subfigure}{0.2465\textwidth}
        \includegraphics[width=0.98\textwidth,clip,trim=0 0 0 0, cfbox=Black 0.5pt 0pt]{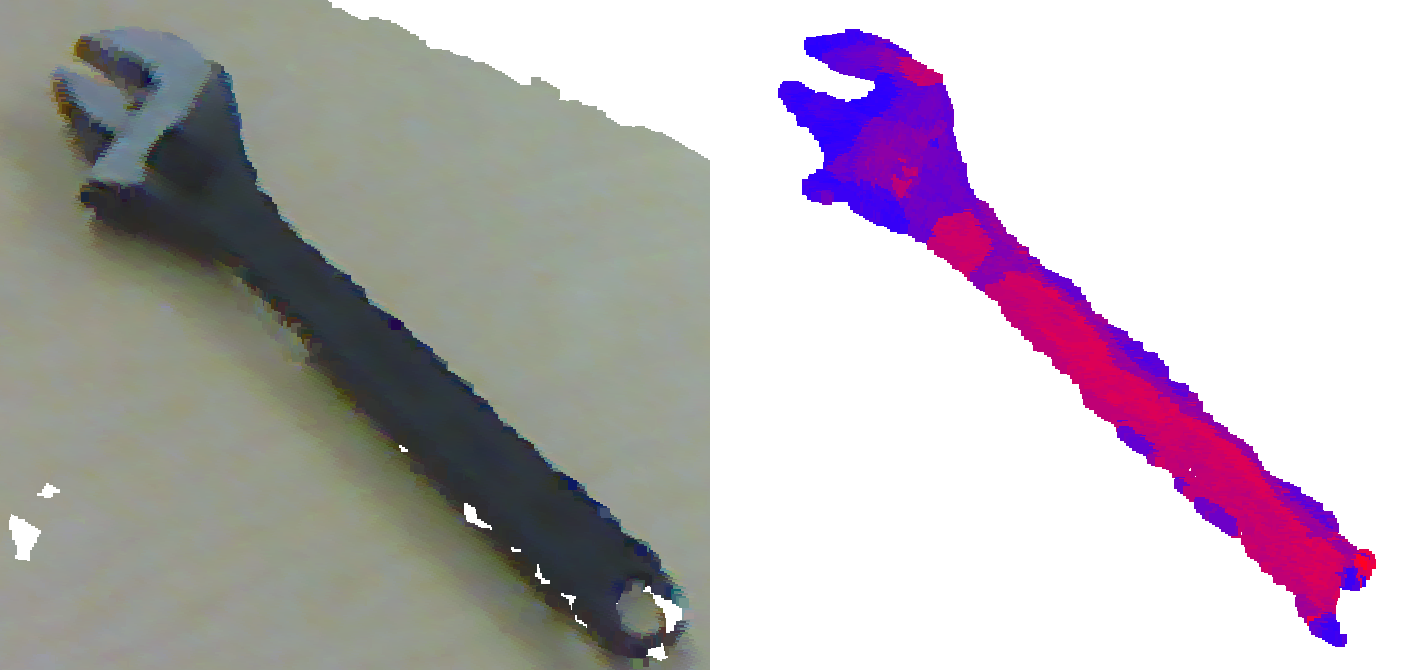}
        \caption{Stirring}\label{fig:feature_projections_stirring3}
    \end{subfigure} &
    \begin{subfigure}{0.2465\textwidth}
        \includegraphics[width=0.98\textwidth,clip,trim=0 0 0 0, cfbox=Black 0.5pt 0pt]{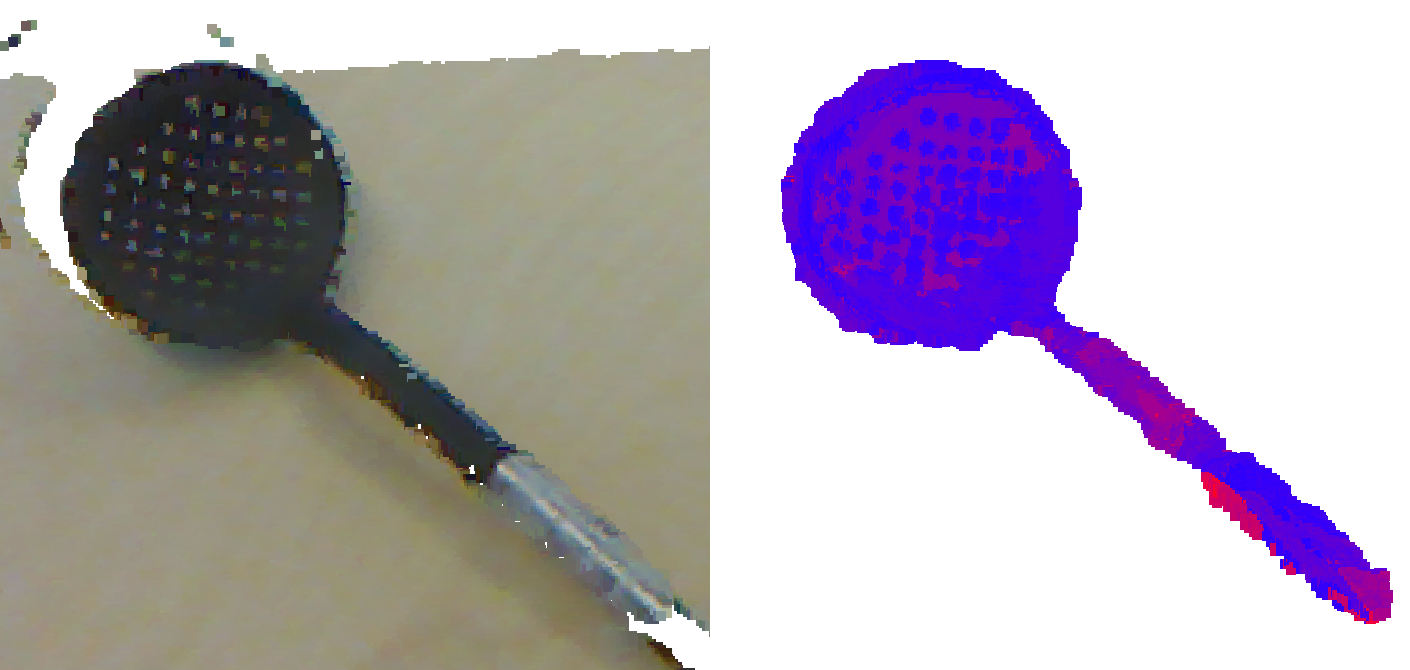}
        \caption{Tool}\label{fig:feature_projections_tool1}
    \end{subfigure} &
    \begin{subfigure}{0.2465\textwidth}
        \includegraphics[width=0.98\textwidth,clip,trim=0 0 0 0, cfbox=Black 0.5pt 0pt]{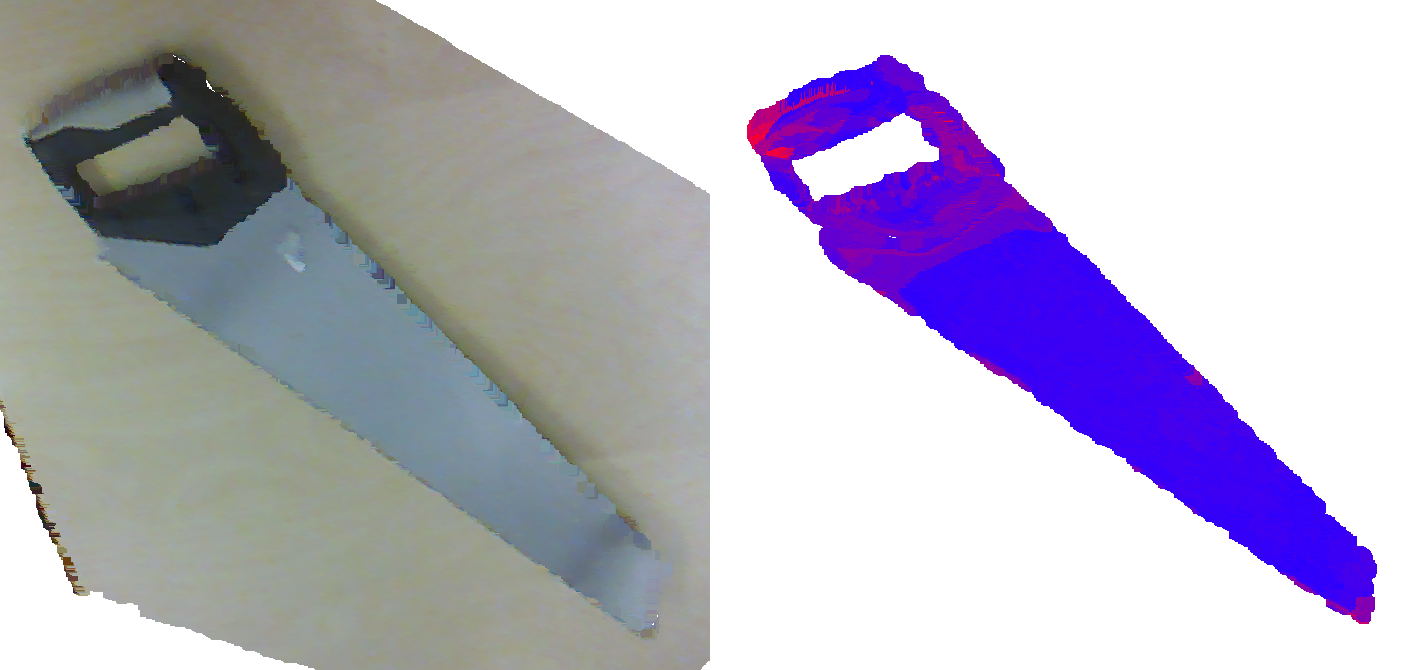}
        \caption{Tool}\label{fig:feature_projections_tool2}
    \end{subfigure} \\[8.ex]
\end{tabular}
\caption{Highlighting of important parts of the object for classifying the object to an affordance according to the feature selection process. The corresponding affordance is given below the image. The bright red parts correspond to important parts and blue to non-important parts. We can see that many of the highlights correspond to parts that humans would agree would be relevant for the specific affordance, even though such a correlation cannot be expected.}
\label{fig:feature_projections}
\end{figure*}

\renewcommand{\thesubfigure}{\alph{subfigure}}

\subsection{Feature Projection}
\label{section:feature_projection}
The second question we set out to answer is: are there certain invariant parts of the objects that are valuable for classifying to an affordance? To investigate this we want to extract the important local features and locate them on the objects that afford the action.

We proceed in the same way as in the feature selection analysis. We take the mean of the magnitudes of $\*L$ over all the runs. From the mean we select the subset of features that are point-cloud based, that is, the gradient, color quantization, FPFH features, and normalize this subset.

To get an indication of the important parts we assign an importance weight value to each point. We compute it by summing, the feature weights associated with the point according to each of the selected features,

\begin{equation}
    c_j = \sum_{f \in Features} w_f[\,f(p_j)\,].
\end{equation}

Here $f$ is a feature function that takes a point cloud index and returns an index corresponding to the weight value for that feature and $w_f$ is the weight vector for the feature $f$. For example, for the BoW FPFH features each codeword has a weight, to find the weight we thus classify a point to a codeword and look up the index for that codeword in the weight vector.

To color the object we divide all the values by the max value taken over all points. We input the values to a gradient function between red and blue, such that values close to the max value becomes red, and values close to zero becomes blue.

Before we analyze the results in Fig.\ref{fig:feature_projections} we want to bring up one important point that cannot be stressed enough: humans and robots are different sensorimotor systems. We have different feature representations and mechanisms for detecting invariant features. Therefore, we cannot expect the invariant selected parts of the objects to be the same for robots and humans. Our approach might detect invariant features that humans are unable to detect or understand. The important part is the consistency in the invariances across objects. With that being said it would be interesting if there is a correspondence between the invariant parts selected by our model and what one can expect from a human.

The objects selected in Fig.\ref{fig:feature_projections} is just a small subset of all positive examples, roughly 1400, but gives a good representation of the main results. 

For \textbf{Drinking}, Fig.\ref{fig:feature_projections_drinking1}-\ref{fig:feature_projections_drinking2}, the highlighted part is the rounded back part of the object. The back part was selected in a similar fashion across most of the objects even for such diverse objects such as the smaller bowl and the teapot.

For \textbf{Eating From}, Fig.\ref{fig:feature_projections_eatingfrom1}-\ref{fig:feature_projections_eatingfrom3}, we see that the algorithm highlighted the flat bottom for two of the objects in  Fig.\ref{fig:feature_projections_eatingfrom1} and Fig.\ref{fig:feature_projections_eatingfrom3} but not in Fig.\ref{fig:feature_projections_eatingfrom2}. This highlights the difficulty in generalizing from a couple of highlights. What these three images show is that the flat parts are important for categorizing those two objects while the sides of the frying pan in Fig.\ref{fig:feature_projections_eatingfrom2} is more important than its flat part for categorizing to the affordance. Despite this, a majority of the objects in the category shows highlighting of the flat or base parts.

\textbf{Handle Grasping}, Fig.\ref{fig:feature_projections_handlegrasping1}-\ref{fig:feature_projections_handlegrasping3}, gave mixed results. Many objects had colorings similar to those in  Fig.\ref{fig:feature_projections_handlegrasping1}-\ref{fig:feature_projections_handlegrasping2}. However, we also had a number of objects where the algorithm either selected the whole object or the connecting part where the handle meets the tool part as in Fig.\ref{fig:feature_projections_handlegrasping3}. We expected this as the connecting part is a common shape across objects with handles.

In \textbf{Hanging}, Fig.\ref{fig:feature_projections_hanging1}-\ref{fig:feature_projections_hanging2}, we gave the algorithm a number of objects with loops. The results were not satisfactory. On one hand, we had results as in Fig.\ref{fig:feature_projections_hanging1}, yet most results were similar to Fig.\ref{fig:feature_projections_hanging2} with significant noise. A closer inspection revealed that a large number of cups skewed the results towards detecting cylindrical parts. The set of objects affording \textbf{Loop Grasping}, a subset of hanging, Fig.\ref{fig:feature_projections_loopgrasping1}-\ref{fig:feature_projections_loopgrasping2}, showed similar effects. 

The \textbf{Lifting Top}, Fig.\ref{fig:feature_projections_liftingtop1}-\ref{fig:feature_projections_liftingtop2}, also gave mixed results. The objects varied significantly in shape and we expected the algorithm to detect the small correlations across the objects given by the shape of the tops. The results show to the contrary that detecting small shapes is difficult at best due to the Kinect's low resolution and level of noise.

\textbf{Opening}, Fig.\ref{fig:feature_projections_opening1}-\ref{fig:feature_projections_opening3}, were perhaps the most surprising results. The objects had large variations in shape, ranging from toothpaste tubes to milk-cartons and bottles. We, therefore, considered it to be one of the more difficult categories.  Despite this, the algorithm consistently highlighted parts of the objects approached for opening for a majority of the objects.

For \textbf{Rolling}, Fig.\ref{fig:feature_projections_rolling1}-\ref{fig:feature_projections_rolling2}, we expected results where the whole object was colored. This happened in the majority of the objects, but there was also some with spurious colorings such as in Fig.\ref{fig:feature_projections_rolling2} where the results were more difficult to interpret.

\textbf{Stacking}, Fig.\ref{fig:feature_projections_stacking1}-\ref{fig:feature_projections_stacking3}, proved to be quite a good illustration of the point made in the beginning about difference in sensorimotor systems. We expected a coloring of the flat parts, but what actually is the common denominator are the edges. The algorithm selected edges similar to those for a majority of the objects.

Finally \textbf{Stirring}, Fig.\ref{fig:feature_projections_stirring1}-\ref{fig:feature_projections_stirring3} and \textbf{Tool}, Fig.\ref{fig:feature_projections_tool1}-\ref{fig:feature_projections_tool1}, gave very interesting results. The objects contained in these two categories are similar and as we can see from Fig.\ref{fig:feature_projections_stirring1}-\ref{fig:feature_projections_stirring3} the algorithm has selected the whole handle part with almost uncanny certainty. Seemingly the algorithm has picked up the rule that objects that afford stirring should have thin and elongated handle parts.

To conclude, the above results show good consistency in selecting sensible parts of the objects in most of the categories. It is clear that we need more data points for the results that showed low consistency such as in *Hanging* and *Loop Grasping*. For example, the algorithm will benefit from more negative examples such as cups without or occluded handles. Creating good datasets with sensible labelings for learning complex abstractions is a trial and error process since the features that you expect to be important might not be. Further on, better depth resolution with less noise will provide a major improvement. For example, flat surfaces are not always interpreted as flat due to the noise. This makes the FPFH BoW features map flat surfaces differently thus introducing large variance in shapes that might not be that different. Lastly, the analysis we made of the selected features differed, in some categories significantly, from the analysis of the projected features. This shows, as mentioned earlier, that drawing conclusions from the belief that different sensorimotor systems will produce similar results can be precarious. 

\begin{table}
    \resizebox{0.99\linewidth}{!}{%
    \def\arraystretch{1.25}%
    \begin{tabular}{l l l l }
    \textbf{Affordance} & \textbf{1.} & \textbf{2.} & \textbf{3.} \\ \arrayrulecolor{safegray}\hline
    \rowcolor{gainsboro}
    \textbf{Containing} & Shaking & Opening & Squeezing\\
    \textbf{Cutting} & Tool & Hammering & Stirring\\
    \rowcolor{gainsboro}
    \textbf{Drinking} & Putting & Loop Grasping & Tool\\
    \textbf{Eating From} & Putting & Loop Grasping & Drinking\\
    \rowcolor{gainsboro}
    \textbf{Hammering} & Tool & Stirring & Scraping\\
    \textbf{Handle Grasping} & Tool & Scraping & Stirring\\
    \rowcolor{gainsboro}
    \textbf{Hanging} & Loop Grasping & Drinking & Spraying\\
    \textbf{Lifting Top} & Opening & Squeezing & Containing\\
    \rowcolor{gainsboro}
    \textbf{Loop Grasping} & Hanging & Drinking & Rolling\\
    \textbf{Opening} & Containing & Shaking & Lifting Top\\
    \rowcolor{gainsboro}
    \textbf{Playing} & Pounding & Shaking & Spraying\\
    \textbf{Pounding} & Rolling & Squeezing & Drinking\\
    \rowcolor{gainsboro}
    \textbf{Pouring} & Shaking & Containing & Opening\\
    \textbf{Putting} & Drinking & Loop Grasping & Hanging\\
    \rowcolor{gainsboro}
    \textbf{Rolling} & Pounding & Lifting Top & Squeezing\\
    \textbf{Scraping} & Tool & Stirring & Handle Grasping\\
    \rowcolor{gainsboro}
    \textbf{Shaking} & Containing & Opening & Squeezing\\
    \textbf{Spraying} & Hanging & Squeezing Out & Squeezing\\
    \rowcolor{gainsboro}
    \textbf{Squeezing} & Containing & Shaking & Lifting Top\\
    \textbf{Squeezing Out} & Spraying & Squeezing & Lifting Top\\
    \rowcolor{gainsboro}
    \textbf{Stacking} & Lifting Top & Squeezing & Putting\\
    \textbf{Stirring} & Scraping & Tool & Hammering\\
    \rowcolor{gainsboro}
    \textbf{Tool} & Handle Grasping & Scraping & Stirring\\ \arrayrulecolor{safegray}\hline
    \end{tabular}
    }
    \caption{The three nearest neighbors for each affordance. We compute the distances using the KL-divergence between the Gaussian distributions over the magnitude vectors of the affordance transforms, $\*L$. Distances are therefore non-symmetric.}
    \label{fig::transformag_distances}
\end{table}

\subsection{Affordance Association}
Finally, we examine how the different affordances relate to each other. We start by assuming that the magnitude of $\*L$ has a multivariate Gaussian distribution. We compute the mean and covariance by treating all the 25 runs as samples from the distribution. We can now measure the similarity between the affordances using the KL-divergence.

In Table \ref{fig::transformag_distances} we list the 3 nearest neighbors (NN) for each affordance. Since the KL-divergence is asymmetrical the NN of one affordance might not be the NN of the other.

From Table \ref{fig::transformag_distances} we can see that most of the affordances that we expected to be close to each other are in fact close. For example, objects that afford tool use are similar to objects that afford handle grasping, scraping, and stirring. Rolling is close to Lifting Top and Squeezing, Loop Grasping is close to Hanging and Drinking, and Cutting is close to Tools. Stacking is close to objects that affords Lifting Top and Putting, etc. The results clearly show that our approach can learn to relate affordances in a consistent and sensible manner. 

One interpretation of the KL-divergence is the amount of information one learns of the true distribution from the information given by another distribution. In our context, this means, how much an affordance says about the features that are important for another affordance. Learning to associate affordances implies learning the interrelation between similar affordances and the objects that make up the clusters of association. This deeper understanding is key to generalizing and abstracting affordances. Practically, this knowledge has the potential to help a robot perform an unknown action demonstrated by another actor. It can do this by analyzing the affordances of the object being manipulated and figuring out what features might be important from what it has learned from other objects effectively bootstrapping the learning process.

\section{Conclusion}\label{conclusion}

We started out with the simple notion of distance as a proxy for
similarity. This guided us to learn a transform of the feature space
that put similar items close and dissimilar items far away. Objects are
usually similar in only a few aspects of their representation and we,
therefore, penalized parts of the feature space that were not relevant
for classifying to the affordance.

We analyzed the penalized transform to deduce the relevant features and
provide a grounding of the affordances. Since some of the feature space
was tied to a point cloud representation we could locate important parts
of the objects for classifying to an affordance. Our model is thus proof
of concept that applying a sensible approach to reasoning about
similarity facilitates the ability to learn abstractions of categories
without the need for pixel ground truths, pre-segmentation, other cues,
and heuristics.

Furthermore, we showed that the model can learn to associate categories
with each other. Instead of analyzing the transformed data, as is
common, we analyzed the feature transforms themselves, computing
distances between them. Again using distance as a proxy for similarity.
The key is the realization that the transform itself contains the
information necessary to reason about the category. The learned
similarities between the affordances proved to be sensible and gave
insight into how an agent can learn to reason about categories.

The shortcomings of our model are obvious. Firstly, stacking designed
features is not a viable option for a fully autonomous system, it will
need to learn the features from the data. This implies that future work
should focus on finding ways to analyze and compare activations in deep
nets e.g.~\cite{DBLP:conf/iros/KuLG17}, either by developing retinotopic
feedback loops similar to how human vision works or other recurrent ways
of learning abstractions, however, without the need for pixel-wise
labeling. Further on, when creating these abstractions we need to
understand to what a degree we should mimic human capabilities, as this
will be a crucial component in human-robot interaction.

Secondly, we showed that there is sufficient information in the shape of
objects to ground the affordances. However, for a robot to gain a
complete understanding of an affordance, it will have to interact with
the objects and ground all observed sensorimotor input, both
proprioceptive and exteroceptive. If we want grounding and abstraction
to be as fluent and effortless as in humans, to enable high-level
reasoning, future work needs to focus on building this knowledge in a
holistic fashion.

\bibliographystyle{elsarticle-num-names}
\bibliography{references}

\end{document}